\documentclass[lettersize,journal]{IEEEtran}
\usepackage{amsmath,amsfonts}
\usepackage{orcidlink}

\usepackage[nocompress]{cite}
\hypersetup{
colorlinks,
citecolor=blue,
linkcolor=blue,
urlcolor=blue
}

\usepackage{mathrsfs}
\usepackage{algorithmic}
\usepackage{algorithm}

\usepackage{array}
\usepackage[caption=false,font=normalsize,labelfont=rm,textfont=rm]{subfig}
\usepackage{textcomp}
\usepackage{stfloats}
\usepackage{url}
\usepackage{bm}
\usepackage{verbatim}
\usepackage{graphicx}

\usepackage{tabularray}
\usepackage{siunitx}
\usepackage{graphicx}
\usepackage[normalem]{ulem}
\usepackage{array}

\begin{document}

\title{Superpixel Graph Contrastive Clustering with Semantic-Invariant Augmentations for Hyperspectral Images}

\author{
  Jianhan~Qi$^{\orcidlink{0009-0004-3810-083X}}$,
  Yuheng~Jia$^{\orcidlink{0000-0002-3907-6550}}$,~\IEEEmembership{Member,~IEEE,}
  Hui~Liu$^{\orcidlink{0000-0003-2159-025X}}$,
  Junhui~Hou$^{\orcidlink{0000-0003-3431-2021}}$,~\IEEEmembership{Senior Member,~IEEE}
  \thanks{Manuscript received 9 February 2024; revised 16 May 2024; accepted 20 June 2024. This work was supported in part by the National Natural Science Foundation of China under Grant 62106044, in part by the Natural Science Foundation of Jiangsu Province under Grant BK20210221, and in part by the Hong Kong UGC under grant UGC/FDS11/E02/22. This article was recommended by Associate Editor Z. Tao. \textit{(Corresponding author: Yuheng Jia.)}}
  \thanks{Jianhan Qi is with the School of Software Engineering, Southeast University, Nanjing 210096, China (e-mail: qjh@seu.edu.cn).}
  \thanks{Yuheng Jia is with the School of Computer Science and Engineering, Southeast University, Nanjing 210096, China, and also with the Key Laboratory of New Generation Artificial Intelligence Technology and Its Interdisciplinary Applications (Southeast University), Ministry of Education, China (e-mail: yhjia@seu.edu.cn).}
  \thanks{Hui Liu is with the Department of Computing and Information Sciences, Saint Francis University, Hong Kong, China (e-mail: h2liu@sfu.edu.hk).}
  \thanks{Junhui Hou is with the Department of Computer Science, City University of Hong Kong, Hong Kong, China (e-mail: jh.hou@cityu.edu.hk).}
}

\markboth{Manuscript Submitted to IEEE TCSVT}%
{Shell \MakeLowercase{\textit{et al.}}: A Sample Article Using IEEEtran.cls for IEEE Journals}

\maketitle

\begin{abstract}
  Hyperspectral images (HSI) clustering is an important but challenging task. The state-of-the-art (SOTA) methods usually rely on superpixels, however, they do not fully utilize the spatial and spectral information in HSI 3-D structure, and their optimization targets are not clustering-oriented. In this work, we first use 3-D and 2-D hybrid convolutional neural networks to extract the high-order spatial and spectral features of HSI through pre-training, and then design a superpixel graph contrastive clustering (SPGCC) model to learn discriminative superpixel representations. Reasonable augmented views are crucial for contrastive clustering, and conventional contrastive learning may hurt the cluster structure since different samples are pushed away in the embedding space even if they belong to the same class. In SPGCC, we design two semantic-invariant data augmentations for HSI superpixels: pixel sampling augmentation and model weight augmentation. Then sample-level alignment and clustering-center-level contrast are performed for better intra-class similarity and inter-class dissimilarity of superpixel embeddings. We perform clustering and network optimization alternatively. Experimental results on several HSI datasets verify the advantages of the proposed SPGCC compared to SOTA methods. Our code is available at \url{https://github.com/jhqi/spgcc}.
\end{abstract}


\section{Introduction\label{introduction}}
\IEEEPARstart{H}{yperspectral} remote sensing techniques employ sensors to collect the reflectance of land-cover materials in hundreds of narrow and contiguous spectral bands, generating hyperspectral images (HSI), which have been widely applied in various fields, such as agricultural monitoring\cite{ref1}, mineral exploration\cite{ref2} and urban planning\cite{ref3}. Grouping HSI into several classes is essential in many applications\cite{jigsaw, tcsvt_hsi1, tcsvt_hsi2, tcsvt_hsi3}. Recent state-of-the-art methods have achieved high accuracy in the supervised HSI classification task\cite{reviewer1_cls1, reviewer1_cls2, reviewer1_cls3, reviewer1_cls4}. However, due to the high cost of manual annotating, obtaining pixel-level labels is difficult.

Unsupervised clustering methods\cite{sglsc, ncsc, dgae, lowpass} are gaining increasing attention. These methods aim to mine the relationships between pixels in the absence of labels and assign similar pixels to the same cluster. For example, center-based clustering methods, e.g., K-means\cite{kmeans} and fuzzy c-means\cite{fuzzycmeans}, assume that instances have a spherical cluster structure, which may be not satisfied in the raw feature space of HSI. Subspace-based clustering methods, e.g., sparse subspace clustering\cite{ssc}, low-rank subspace clustering\cite{lrsc}, and dense subspace clustering\cite{densesc}, divide instances into different low-dimensional subspaces, which are susceptible to redundant spectral features. Additionally, considering the large quantity of HSI pixels, subspace clustering methods may lead to unbearable computational costs because they need to calculate the self-represented coefficient matrix with time complexity of $\mathcal{O}(n^{3})$, where $n$ is the number of pixels. Therefore, some competitive subspace-based methods are limited to experimenting on a subset of the complete HSI\cite{ekgcsc,bgp}.

To this end, superpixel segmentation techniques, e.g., ERS\cite{ers} and SLIC\cite{slic}, which divide HSI into irregular superpixels with high homogeneity, have been widely used in HSI clustering. In these methods, pixels within the same superpixel are assumed to belong to the same class, and superpixel-level clustering is performed. As the number of superpixels is much smaller than that of pixels, the computational efficiency can be greatly improved by transforming the clustering task from pixel level to superpixel level. Moreover, superpixel representations are smoother and less susceptible to spectral noise\cite{survey}. Some methods have combined superpixel segmentation with traditional clustering techniques, resulting in competitive performance. For example, Hinojosa \textit{et al.} took advantage of HSI’s neighboring spatial information to improve the superpixel-level subspace clustering algorithm\cite{afast}. Zhao \textit{et al.} introduced both global and local similarity matrices to boost the clustering accuracy\cite{sglsc}.

Since superpixels have a clear spatial adjacent relationship with the neighbor superpixels, it is natural to process superpixels as graph nodes with edges connected to their spatial neighbors, and many efforts have been devoted to extracting structured information of superpixels with graph learning methods, e.g., graph convolutional networks (GCN)\cite{gcn} and graph attention networks (GAT)\cite{gat}. The most popular training framework, graph autoencoder (GAE)\cite{gae}, takes the reconstruction of adjacency matrices as the optimization target and learns low-dimensional representations of graph nodes. Based on GAE, Zhang \textit{et al.} achieved HSI clustering by building spatial and spectral similarity graphs of superpixels and extracting both spatial and spectral features simultaneously with dual graph convolutional encoders\cite{dgae}. Ding \textit{et al.} proposed a low-pass GCN with the layer-wise attention mechanism to extract the smooth and local features of superpixels\cite{lowpass}. Moreover, the auxiliary distribution loss is often introduced to guide the more concentrated cluster structure\cite{dec, tcsvt_graph_clustering}. Generally, such methods model the relationships between superpixels with a similarity matrix and gain superpixel representations by GAE. \textbf{\textit{However, they mainly have two drawbacks.}} First, the input superpixel features are directly represented as the average of internal pixels, ignoring the 3-D structure of HSI at the pixel level, thus not fully utilizing the spatial and spectral information of HSI. Second, the learned superpixel representations of the above methods may be not clustering-friendly due to their training objectives are not clustering-oriented. Specifically, features that are easy to cluster tend to exhibit intra-class similarity and inter-class dissimilarity, but GAE-based methods mainly aim to reconstruct the input graph structure, thus limiting the clustering performance.

To overcome the aforementioned drawbacks, we propose a superpixel-level clustering method for HSI with pixel-level pre-training, named superpixel graph contrastive clustering (SPGCC). First, we exploit the 3-D structure of HSI by dividing the original HSI into 3-D cubes. Then, 3-D and 2-D hybrid convolutions are performed on pixel cubes to extract the high-order spatial and spectral features of HSI simultaneously which can be preserved in superpixel-level representations. Next, we aggregate the topological information of superpixels from neighborhoods by GCN. To learn the discriminative and clustering-friendly superpixel representations, we utilize contrastive learning (CL), which brings samples close to their augmented views in the embedding space while pushing different samples' representations far apart\cite{simclr, moco, byol}. Proper data augmentations are crucial for CL, as augmented views and original samples should share the same semantics. To this end, we design two effective data augmentations for superpixels, i.e., pixel sampling augmentation and model weight augmentation, to obtain semantic-invariant augmented views. Additionally, traditional CL-based clustering methods treat different samples as negative pairs and push their representations to be far apart from each other\cite{cc}, even for samples in the same class, which is harmful to clustering accuracy. Differently, we perform sample-level alignment and clustering-center-level contrast on superpixels to obtain better intra-class similarity and inter-class dissimilarity. To mitigate the impact of outliers, we select samples with high confidence to recompute clustering centers. In the whole framework, we alternatively conduct clustering and network optimization, using clustering results to guide the optimization. The proposed SPGCC has an improvement of 2.5\%, 2.4\%, and 2.7\% over the best compared methods on three different HSI datasets.

The main contributions of this paper are summarized as follows:
\begin{enumerate}
  \item[1)] We perform pixel-level pre-training with 3-D and 2-D hybrid CNN to extract high-order spatial and spectral information of HSI. Moreover, we separate it with the clustering network based on GCN, which enables our method to handle the large-scale HSI clustering task.
  \item[2)] Two types of semantic-invariant data augmentations are proposed for HSI superpixels, which can improve contrastive learning by obtaining reliable positive samples.
  \item[3)] Sample-level alignment and clustering-center-level contrast are combined to address the limitation of traditional contrastive learning methods that treat different samples of the same class as negative pairs, resulting in more clustering-friendly representations.
\end{enumerate}

The rest of this paper is organized as follows. Section \ref{preliminary} briefly reviews preliminary about HSI pixel convolution, graph contrastive learning and HSI data augmentation. Section \ref{proposed_method} presents the proposed SPGCC algorithm in detail, including pixel-level pre-training, graph construction, semantic-invariant superpixel augmentations, and graph contrastive clustering. In Section \ref{result}, we provide experimental results and analyses on several HSI datasets. Finally, we summarize our work in Section \ref{conclusion}.

\section{Preliminary\label{preliminary}}
\subsection{HSI Pixel Convolution}
Convolutional neural networks (CNNs) are powerful tools for extracting high-order features from images. For HSI, a common idea is to expand the pixel into a square neighborhood centered on itself, forming pixel cubes. Conventional 2-D CNNs simply treat the pixel cube as a multi-channel image and perform convolution at each channel, ignoring the rich spectral features of HSI. Therefore, many efforts have been devoted to extracting spatial and spectral features from HSI simultaneously with 3-D CNNs\cite{3dcnn,3dcnn1,3dcnn2}. Considering the high computational cost of 3-D CNNs, Roy \textit{et al.} proposed 3-D and 2-D hybrid CNNs to accelerate computation efficiency\cite{hybridsn}.

Given an HSI cube and a 3-D convolutional kernel, the kernel moves along the width, height, and depth (spectral) dimensions respectively with a fixed step. At each position, the convolution is defined as the Hadamard product of the kernel and the related part of the HSI cube. Similar to 2-D convolution, 3-D convolution is also conducted in a multi-channel way to extract features of different scales. The output value in the $j$-th channel of the $i$-th layer at spatial position $(x, y, z)$, denoted as $v_{i,j}^{x,y,z}$, is calculated by:
\begin{equation}
  \label{3dconv}
  \begin{aligned}
     & v_{i,j}^{x,y,z}=                                                                                                                                                                                                                                     \\
     & \phi(\sum\limits_{\tau=1}^{d_{l-1}} \sum\limits_{\lambda=-\eta}^{\eta} \sum\limits_{\rho=-\gamma}^{\gamma} \sum\limits_{\sigma=-\delta}^{\delta} \omega_{i,j,\tau}^{\sigma,\rho,\lambda} \times v_{i-1,\tau}^{x+\sigma,y+\rho,z+\lambda} + b_{i,j}),
  \end{aligned}
\end{equation}
where $d_{l-1}$ is the number of channels in the $(l-1)$-th layer, $\omega_{i,j,\tau} \in \mathbb{R}^{(2\delta+1)\times(2\gamma+1)\times({2\eta+1})}$ is the kernel's weight parameter in the $j$-th channel of the $i$-th layer related to the $\tau$-th channel of the last layer, superscripts $(\sigma,\rho,\lambda)$ are indices along corresponding dimensions, $b_{i,j}$ is the bias, and $\phi$ is the activation function.

HSI pixel convolution has been widely used in classification tasks, significantly improving the accuracy, especially with 3-D convolution. For clustering tasks, although 2-D convolution has been utilized as the feature extraction network\cite{ncsc, sscc}, 3-D convolution is much less used. An intuitive reason is that 3-D convolution has higher complexity and some clustering networks require a complete batch input, resulting in excessive GPU memory consumption.

\begin{figure*}[ht]
  \centering
  \includegraphics[width=\textwidth]{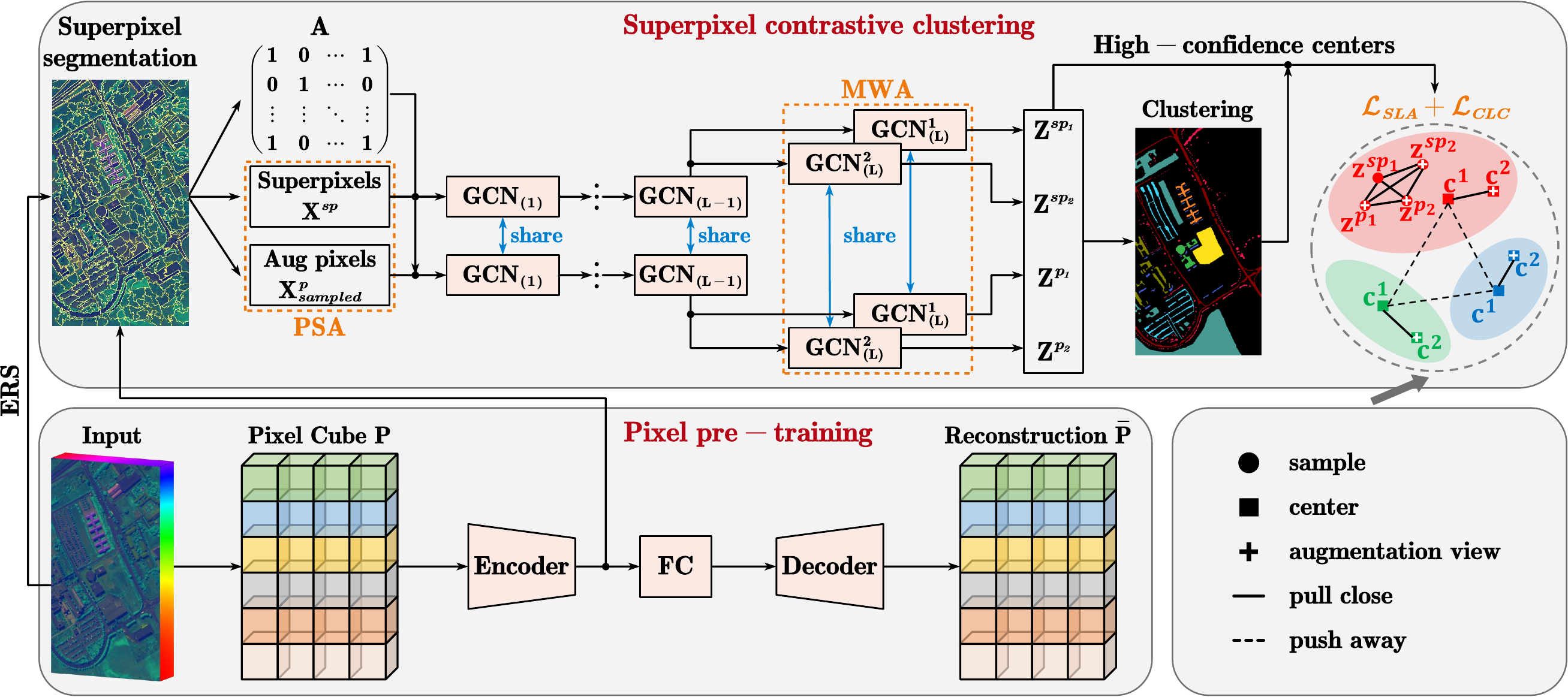}
  \caption{The diagram of superpixel graph contrastive clustering (SPGCC). First, the input HSI is divided into several 3-D pixel cubes, and superpixels are obtained by entropy rate superpixel segmentation (ERS). At pixel-level pre-training, high-order spatial and spectral features are extracted by 3-D and 2-D hybrid CNNs encoder and decoder. At superpixel-level contrastive clustering, model weight augmentation (MWA) and pixel sampling augmentation (PSA) are combined to generate semantic-invariant augmentations, i.e., unshared dual branches in the last layer of graph encoder output different augmented views for superpixels and sampled internal pixels, respectively. Then, K-means is performed to get clustering results, and high-confidence clustering centers of two augmented superpixel views $\mathbf{c}^1, \mathbf{c}^2$ are recomputed based on samples close to the original centers. Finally, the network is trained with sample-level alignment (SLA) and clustering-center-level contrast (CLC), where the different views are encouraged to be similar at both the sample level and clustering center level while the contrast is only performed at the clustering center level. Clustering and network optimization are performed alternatively.}
  \label{pipeline}
\end{figure*}

\subsection{Graph Contrastive Learning}
Several works have applied contrastive learning for graph data due to its remarkable representation ability\cite{simsiam,gcl1,gcl2}. Here, we focus on node-level graph contrastive learning (GCL) and divide it into three parts: graph data augmentation, graph encoder network, and contrastive loss function. A graph is denoted as $g=(\mathcal{V},\mathcal{E})$ where $\mathcal{V}$ is the set of $N$ nodes and $\mathcal{E}$ is the set of $M$ edges. The graph structure is represented by an adjacency matrix $\mathbf{A} \in \mathbb{R}^{N \times N}$ and node attributes are denoted as $\mathbf{X} \in \mathbb{R}^{N \times d}$. We use $\Gamma$ to denote a set of different graph data augmentations and then augmented views are defined as:
\begin{equation}
  \label{graph_data_aug}
  \mathbf{A}^k, \mathbf{X}^k=\gamma^k(\mathbf{A}, \mathbf{X});\gamma^k \in \Gamma.
\end{equation}
Node embeddings are obtained through graph encoder network $f(\cdot)$, e.g., graph convolutional network\cite{gcn} or graph attention network\cite{gat}.

Given two different views of graph nodes, their embeddings are calculated as $\mathbf{Z}^1=f(\mathbf{A}^1, \mathbf{X}^1), \mathbf{Z}^2=f(\mathbf{A}^2, \mathbf{X}^2)$ and $\mathbf{z}_i^k (k=1,2)$ is the $i$-th node's embedding from the $k$-th view. Most GCL methods follow the InfoMax\cite{graph_infomax} principle, which maximizes the mutual information between two augmented views of the graph. According to the most commonly used InfoNCE loss\cite{moco}, embeddings of the same node from different views form positive sample pairs, while others form negative sample pairs. Finally, pairwise objective for $(\mathbf{z}_i^1, \mathbf{z}_i^2)$ is defined as follows:
\begin{equation}
  \label{pairwise_cl_loss}
  \mathcal{L}(\mathbf{z}_i^1, \mathbf{z}_i^2)=-log\frac{exp(sim(\mathbf{z}_i^1, \mathbf{z}_i^2)/\tau)}
  {exp(sim(\mathbf{z}_i^1, \mathbf{z}_i^2)/\tau)+Neg}
\end{equation}
and $Neg$ is defined as
\begin{equation}
  \label{pairwise_neg}
  Neg=\sum_{j\neq i}exp(sim(\mathbf{z}_i^1, \mathbf{z}_j^2)/\tau),
\end{equation}
where $sim(\cdot)$ is the similarity function and $\tau$ is the temperature scaling parameter.

\subsection{Data Augmentation for HSI}
Data augmentation can enhance the quality of original samples or generate additional reliable samples to improve training, which has been widely used in HSI tasks \cite{reviewer2_1, reviewer2_2, sscc}. Here we mainly focus on the impact of data augmentation on superpixel-level graph contrastive learning.

For graph-structured superpixel data with node attributes and the adjacency matrix, general graph data augmentations can be directly applied, such as node mask, node noise, node shuffling, edge perturbation \cite{gcl_survey, Data_Augmentation_on_Graphs}. However, these methods mainly rely on random perturbations of graph nodes or edges, which are uncontrollable and may erroneously alter semantic information of the graph. In contrastive learning, the augmented views are supposed to share the same semantic information with the original samples \cite{moco,simclr}, so the model can learn the common latent features by reducing the distance between original samples and corresponding augmented views in the representation space. Improper data augmentations may lead to negative representation learning. To this end, semantic-invariant data augmentations are crucial for conducting contrastive learning on HSI, which have been rarely investigated.

\section{Proposed Method\label{proposed_method}}
Original HSI is 3-D structured and can be denoted as $\mathcal{S}\in\mathbb{R}^{H\times{W}\times{C}}$ with resolution of $H\times{W}$ and $C$ spectral bands. The target is to assign the total $N$ pixels into $K$ classes where $N=H\times{W}$. Our method consists of pixel-level pre-training and superpixel-level contrastive clustering. Since clustering is performed for superpixels, pixels' labels are inherited from the belonged superpixels. The overall architecture of the proposed method is shown in Fig. \ref{pipeline}. First, we employ a variational autoencoder composed of hybrid 3-D and 2-D CNNs for pixel-level pre-training, extracting high-order spatial and spectral features from HSI pixels. Subsequently, superpixel segmentation is performed and the graph structure of superpixels is constructed. In superpixel graph contrastive clustering, we propose two semantical-invariant data augmentations for superpixels and employ sample-level alignment together with clustering-center-level contrast as the target of network optimization. Clustering results are obtained by K-means and then high-confidence clustering centers are recomputed. We conduct clustering and network optimization alternatively.
\begin{table}[ht]
  \caption{Architecture of Pixel Pre-training Network and -1 Denotes the Batch Size.\label{vae_structure}}
  \centering
  \resizebox{\linewidth}{!}{
    \begin{tblr}{
      cells = {c},
      cell{1}{1} = {c=3}{},
      cell{1}{4} = {c=3}{},
      vline{2} = {1}{0.03em},
      vline{4} = {-}{0.03em},
      hline{1,17} = {-}{0.08em},
          hline{2-3} = {-}{0.03em},
        }
      Encoder &              &                    & Decoder  &               &                    \\
      Layer   & Kernel       & Output             & Layer    & Kernel        & Output             \\
      Input   & -            & {[}-1,1,30,27,27]  & Input    & -             & {[}-1,128]         \\
      Conv3D  & {[}8,7,3,3]  & {[}-1,8,24,25,25]  & FC       & {[}128,256]   & {[}-1,256]         \\
      BN3D    & -            & -                  & FC       & {[}256,23104] & {[}-1,23104]       \\
      Conv3D  & {[}16,5,3,3] & {[}-1,16,20,23,23] & Reshape  & -             & {[}-1,64,19,19]    \\
      BN3D    & -            & -                  & DeConv2D & {[}576,3,3]   & {[}-1,576,21,21]   \\
      Conv3D  & {[}32,3,3,3] & {[}-1,32,18,21,21] & BN2D     & -             & -                  \\
      BN3D    & -            & -                  & Reshape  & -             & {[}-1,32,18,21,21] \\
      Reshape & -            & {[}-1,576,21,21]   & DeConv3D & {[}16,3,3,3]  & {[}-1,16,20,23,23] \\
      Conv2D  & {[}64,3,3]   & {[}-1,64,19,19]    & BN3D     & -             & -                  \\
      BN2D    & -            & -                  & DeConv3D & {[}8,5,3,3]   & {[}-1,8,24,25,25]  \\
      GAP     & -            & {[}-1,64,4,4]      & BN3D     & -             & -                  \\
      Reshape & -            & {[}-1,1024]        & DeConv3D & {[}1,7,3,3]   & {[}-1,1,30,27,27]  \\
      FC      & {[}1024,512] & {[}-1,512]         & BN3D     & -             & -                  \\
      FC      & {[}512,128]  & {[}-1,128]         &          &               &
    \end{tblr}}
\end{table}
\subsection{Pixel-level Pre-training}
In this section, we perform pixel-level pre-training to fully extract spatial and spectral features of HSI. Following the existing structure\cite{hybridsn, contrastnet}, we utilize 3-D and 2-D hybrid CNNs to construct variational autoencoder (VAE)\cite{vae} and train the network without supervision. First, PCA is used to reduce the number of bands for the reduction of computational cost. Then, we use a sliding window to split the HSI into pixel cubes  $\mathbf{P}_i\in\mathbb{R}^{w\times{w}\times{h}}(i=1,2,...,N)$ where $\mathbf{P}_i$ is the $i$-th pixel cube, $w$ and $h$ are the window size and the number of reduced bands respectively. In this manner, pixels are represented by a fixed-size neighborhood, which utilizes spatial information to mitigate spectral variability. The encoder mainly consists of two layers of 3-D convolution and one layer of 2-D convolution. We add a global average pooling layer (GAP) at the end to fix the output dimension. The output of GAP for the $i$-th pixel cube is then reshaped into $\mathbf{x}_i^p \in \mathbb{R}^d$. Next, we use fully-connected layers (FC) to fit the mean vector $\bm{\mu}$ and standard deviation vector $\bm{\sigma}$ of the data distribution, and the latent code $\mathbf{q}$ is calculated with re-parameterization trick:
\begin{equation}
  \label{vae_latent}
  \mathbf{q}=\bm{\mu}+\epsilon \times \bm{\sigma},
\end{equation}
where $\epsilon$ is sampled from standard normal distribution and $\bm{\mu},\bm{\sigma},\mathbf{q}\in \mathbb{R}^{d'}$. The decoder consists of symmetric deconvolution layers to reconstruct the input pixel cube, and its input and output are $\mathbf{q}$ and $\overline{\mathbf{P}}_i$. In addition, we add batch normalization layers after each convolution layer to reduce the risk of overfitting. The specific structure of the pre-training network is shown in Table \ref{vae_structure}.

Loss function of VAE is composed of distribution loss $\mathcal{L}_{\bm{\mu},\bm{\sigma}^2}$ and reconstruction loss $\mathcal{L}_{recon}$, which are calculated by (\ref{vae_dist_loss}) and (\ref{vae_recon_loss}):
\begin{equation}
  \label{vae_dist_loss}
  \mathcal{L}_{\bm{\mu},\bm{\sigma}^2}=\frac{1}{2}\sum_{i=1}^{d'}(\mu_{(i)}^{2}+\sigma_{(i)}^{2}-log\sigma_{(i)}^{2}-1),
\end{equation}
\begin{equation}
  \label{vae_recon_loss}
  \mathcal{L}_{recon}=\frac{1}{2}\sum_{i=1}^{N}(\Vert \mathbf{P}_i-\overline{\mathbf{P}}_i \Vert_F^2),
\end{equation}
where the subscript $(i)$ denotes the $i$-th element in the vector, and the total loss of pre-training is calculated by (\ref{vae_loss}):
\begin{equation}
  \label{vae_loss}
  \mathcal{L}_{pre-train}=\mathcal{L}_{\bm{\mu},\bm{\sigma}^2}+\mathcal{L}_{recon}.
\end{equation}

After pre-training, we save the output of GAP $\mathbf{X}^p=[\mathbf{x}_1^p;\mathbf{x}_2^p;...;\mathbf{x}_N^p]\in \mathbb{R}^{N\times d}$ so that pixel's representations with high-order spatial and spectral information are accessible.

\subsection{Superpixel Segmentation and Graph Construction}
The large number of pixels in HSI results in high computational complexity. Benefiting from the superpixel segmentation technology, HSI can be divided into several homogeneous regions. Specifically, we adopt entropy rate superpixel segmentation (ERS) to divide HSI into $M$ non-overlapping superpixels, denoted as $\mathcal{S}=\bigcup_{i=1}^{M}\mathbf{x}^{sp}_i$ where $\mathbf{x}^{sp}_i$ represents the $i$-th superpixel. As $M \ll N$, the computational complexity can be greatly reduced by converting the clustering task from the pixel level to the superpixel level.

To obtain superpixels' representations, we first construct a map matrix
$\mathbf{T}\in \mathbb{R}^{M\times N}$ where $\mathbf{T}_{i,j}=1$ if the $j$-pixel belongs to the $i$-th superpixel, otherwise 0.

Then each superpixel is represented as the average of pre-trained representations of internal pixels by:
\begin{equation}
  \label{sp_representation}
  \mathbf{X}^{sp}=\widetilde{\mathbf{T}}\mathbf{X}^p,
\end{equation}
where $\mathbf{X}^{sp} \in \mathbb{R}^{M \times d}$ and $\widetilde{\mathbf{T}}$ is row-normalized $\mathbf{T}$. Similarity graph $\mathbf{A} \in \mathbb{R}^{M \times M}$ is constructed based on the spatial adjacent relationship between superpixels, i.e., $a_{m,n}=a_{n,m}=1$ if and only if there exists at least one pair of $(\mathbf{x}^p_i, \mathbf{x}^p_j)$ that is adjacent in HSI where $\mathbf{x}^p_i$ belongs to the $m$-th superpixel and $\mathbf{x}^p_j$ belongs to the $n$-th superpixel.

For simplicity, we use $L$-layers graph convolutional network (GCN) as the graph encoder to aggregate neighborhood information. The re-normalization trick is applied by adding the self-loop:
\begin{equation}
  \label{renorm_adj}
  \widetilde{\mathbf{A}}=\mathbf{A}+\mathbf{I},
\end{equation}
where $\mathbf{I}$ is an identity matrix. The renormalized degree matrix $\widetilde{\mathbf{D}}$ is calculated by $\widetilde{d}_{ii}=\sum_{j}\widetilde{a}_{ij}$. Graph convolution is defined as:
\begin{equation}
  \label{gcn}
  \mathbf{H}_{(l)}=RELU(\widetilde{\mathbf{D}}^{-\frac{1}{2}}\widetilde{\mathbf{A}}\widetilde{\mathbf{D}}^{-\frac{1}{2}}\mathbf{H}_{(l-1)}\mathbf{W}_{(l)}),
\end{equation}
where $\mathbf{H}_{(l)}$ and $\mathbf{W}_{(l)}$ are the output and parameters of the $l$-th layer and $\mathbf{H}_{(0)}=\mathbf{X}^{sp}$, $RELU(\cdot)$ is non-linear activation function. We also denote the last layer's output $\mathbf{H}_{(L)}$ as $\mathbf{Z}$ for convenience.

\subsection{Superpixel Semantic-Invariant Data Augmentations}
Here we propose two semantic-invariant augmentations for superpixels in detail. First, we define pixel sampling augmentation (PSA). Each superpixel contains several internal pixels, and its representation is the average of all these pixels. It is a natural idea that pixels within a superpixel can serve as semantic-invariant augmented views of this superpixel. Thanks to the high precision of superpixel segmentation, internal pixels mostly belong to the same class, and may not lead to semantic change (detailed information could be found in Section \ref{sp_num_acc}). In practice, we randomly sample one pixel from each superpixel as its positive augmentation view. All the sampled pixels are denoted as $\mathbf{X}^p_{sampled} \in \mathbb{R}^{M \times d}$ which has the same shape as $\mathbf{X}^{sp}$.

The second superpixel augmentation, model weight augmentation (MWA), is enlightened by the multi-head idea\cite{transformer, CCGC}. We design dual branches in the final layer of the GCN encoder to obtain augmented views with semantic consistency. Specifically, the $L$-th layer contains two branches with unshared parameter matrices $\mathbf{W}_{(L)}^1$ and $\mathbf{W}_{(L)}^2$ which have the same architecture. Due to the different random initialization, two branches' outputs $\mathbf{H}_{(L)}^1$ and $\mathbf{H}_{(L)}^2$ represent two different views of the input superpixel but with the same semantic, i.e., a further aggregation result based on the $(L-1)$-th layer. Considering the parameter efficiency and the diversity of positive samples, parameters are only unshared in the final layer, while shared in previous layers.

Combining PSA with MWA, we can obtain different augmented views of the superpixel. Specifically, representations of superpixels $\mathbf{X}^{sp}$ and the corresponding sampled pixels $\mathbf{X}^p_{sampled}$ are the inputs of the graph encoder. In the first $L-1$ layers, embeddings of superpixels and sampled pixels are calculated by Eq. (\ref{gcn}), denoted as $\mathbf{H}_{(L-1)}^{sp}$ and $\mathbf{H}_{(L-1)}^{p}$ respectively. In the last layer, $\mathbf{H}_{(L-1)}^{sp}$ and $\mathbf{H}_{(L-1)}^{p}$ are inputted into both two branches, thus the outputs include four embeddings: $\mathbf{Z}^{sp_1}, \mathbf{Z}^{sp_2}, \mathbf{Z}^{p_1}, \mathbf{Z}^{p_2}$ where $\mathbf{Z}^{sp_1}$ and $\mathbf{Z}^{p_1}$ are the outputs of the first branch for superpixels and sampled pixels, while $\mathbf{Z}^{sp_2}$ and $\mathbf{Z}^{p_2}$ are of the second. Subsequently, we normalize each row of these embeddings with $l_2$-norm.

\begin{algorithm}[ht]
  \caption{SPGCC}
  \label{spgcc_alg}
  \begin{algorithmic}[1]
    \REQUIRE HSI $\mathcal{S}$; Superpixel segmentation $\mathbf{T}$; Max iteration $I$.\\
    \ENSURE Clustering results $\mathbf{R}$.\\
    \STATE Split the input HSI $\mathcal{S}$ into pixel cubes $\mathbf{P}$ with a sliding window.
    \STATE Perform pixel-level pre-training on $\mathbf{P}$ to obtain pixel representations $\mathbf{X}^p$.
    \STATE Calculate the input superpixel features $\mathbf{X}^{sp}$ with (\ref{sp_representation}) and get adjacency matrix $\mathbf{A}$.
    \FOR{$i=1$ to $I$}
    \STATE Randomly sample one pixel from each superpixel to get $\mathbf{X}^{p}_{sampled}$.
    \STATE Encode representations of superpixels and internal sampled pixels with (\ref{gcn}) to obtain $\mathbf{Z}^{sp_1}, \mathbf{Z}^{sp_2}, \mathbf{Z}^{p_1}, \mathbf{Z}^{p_2}$.
    \STATE Concatenate $\mathbf{Z}^{sp_1}$ and $\mathbf{Z}^{sp_2}$ to get $\mathbf{Z}^{sp}$.
    \STATE Perform K-means on $\mathbf{Z}^{sp}$ to get clustering results $\mathbf{R}$ and centers $\mu_k$.
    \STATE Select high-confidence samples according to (\ref{dis}).
    \STATE Recompute high-confidence centers with (\ref{center}).
    \STATE Calculate sample-level alignment loss $\mathcal{L}_{SLA}$ with (\ref{loss_sla}).
    \STATE Calculate clustering-center-level contrast loss $\mathcal{L}_{CLC}$ with (\ref{loss_clc}).
    \STATE Calculate the total loss $\mathcal{L}$ with (\ref{total_loss}).
    \STATE Optimize model parameters with Adam.
    \ENDFOR
    \STATE Map clustering results $\mathbf{R}$ to superpixel level to pixel level.
    \RETURN $\mathbf{R}$
  \end{algorithmic}
\end{algorithm}

\subsection{Superpixel Contrastive Clustering}
In this section, we jointly get clustering results and optimize superpixels' embeddings under a contrastive learning framework. First, we concatenate the two views $\mathbf{Z}^{sp_1}$ and $\mathbf{Z}^{sp_2}$ to get the final embeddings of superpixels:
\begin{equation}
  \label{zsp}
  \mathbf{Z}^{sp}=[\mathbf{Z}^{sp_1}, \mathbf{Z}^{sp_2}].
\end{equation}
Then we perform K-means on $\mathbf{Z}^{sp}$ to obtain clustering assignments and the $k$-th clustering center is denoted as $\bm{\mu}_k (k=1,2,...,K)$. Considering the clustering centers are susceptible to outliers, to select high-confidence samples and extract more reliable clustering information\cite{CCGC}, we calculate the distance between samples and their nearest clustering centers:
\begin{equation}
  \label{dis}
  d_i=\min_{1 \leq k \leq K} \Vert \mathbf{z}^{sp}_i-\bm{\mu}_k \Vert_2^2,
\end{equation}
where $\mathbf{z}^{sp}_i$ and $d_i$ denote the $i$-th superpixel embedding and its distance to the clustering center. Smaller distance means higher clustering confidence, so superpixels are sorted in ascending order according to their distances, and the top $\lambda$ are selected as high-confidence samples. Indices of high-confidence samples in class $k$ are denoted as $h_k=\{h_k^1,h_k^2,...,h_k^{n_k}\}$. Next, high-confidence clustering centers of two views $\mathbf{c}_k^1, \mathbf{c}_k^2(k=1,2,...,K)$ that are more reliable than $\bm{\mu}_k$ are recomputed only based on these high-confidence samples:
\begin{equation}
  \label{center}
  \mathbf{c}_k^1=\frac{1}{|h_k|}\sum_{i\in h_k}\mathbf{z}_i^{sp_1}, \mathbf{c}_k^2=\frac{1}{|h_k|}\sum_{i\in h_k}\mathbf{z}_i^{sp_2}.
\end{equation}

Traditional contrastive clustering methods suffer from low intra-class consistency caused by wrong negative pairs\cite{propos}. To overcome this deficiency, we utilize sample-level alignment (SLA) and clustering-center-level contrast (CLC) as optimization objectives. Briefly, at the sample level, we do not treat different superpixels as negative pairs since they may belong to the same class. Superpixel embeddings are only encouraged to be aligned with their semantic-invariant augmented views. At the clustering center level, for a certain clustering center, it is natural to set the other $K-1$ centers as its negative views and perform contrast. Combining SLA with CLC, we hope to get superpixel embeddings with better intra-class similarity and inter-class dissimilarity.

For SLA, since we have obtained multiple augmented views of superpixels, we perform six pairs of alignments in total, which can be divided into three groups. The first group, $\mathbf{Z}^{sp_1}$ and $\mathbf{Z}^{sp_2}$, $\mathbf{Z}^{p_1}$ and $\mathbf{Z}^{p_2}$, aims to encourage the outputs of dual branches to be consistent. The second group, $\mathbf{Z}^{sp_1}$ and $\mathbf{Z}^{p_1}$, $\mathbf{Z}^{sp_2}$ and $\mathbf{Z}^{p_2}$, aims to pull the embeddings of superpixels close to sampled internal pixels. The third group, $\mathbf{Z}^{sp_1}$ and $\mathbf{Z}^{p_2}$, $\mathbf{Z}^{sp_2}$ and $\mathbf{Z}^{p_1}$, combines the above two targets together. Loss functions for three groups of alignments are defined as:
\begin{equation}
  \label{group1}
  \mathcal{L}_{SLA}^{group_1}=\Vert \mathbf{Z}^{sp_1}-\mathbf{Z}^{sp_2} \Vert_F^2 + \Vert \mathbf{Z}^{p_1}-\mathbf{Z}^{p_2} \Vert_F^2,
\end{equation}
\begin{equation}
  \label{group2}
  \mathcal{L}_{SLA}^{group_2}=\Vert \mathbf{Z}^{sp_1}-\mathbf{Z}^{p_1} \Vert_F^2 + \Vert \mathbf{Z}^{sp_2}-\mathbf{Z}^{p_2} \Vert_F^2,
\end{equation}
\begin{equation}
  \label{group3}
  \mathcal{L}_{SLA}^{group_3}=\Vert \mathbf{Z}^{sp_1}-\mathbf{Z}^{p_2} \Vert_F^2 + \Vert \mathbf{Z}^{sp_2}-\mathbf{Z}^{p_1} \Vert_F^2,
\end{equation}
and the total loss of SLA is:
\begin{equation}
  \label{loss_sla}
  \mathcal{L}_{SLA}=\frac{1}{6}(\mathcal{L}_{SLA}^{group_1}+\mathcal{L}_{SLA}^{group_2}+\mathcal{L}_{SLA}^{group_3}).
\end{equation}

For CLC, we push away embeddings of different clustering centers, while pulling together the same clustering center's different views:
\begin{equation}
  \label{loss_clc}
  \mathcal{L}_{CLC}=\frac{1}{K}\sum_{k=1}^{K}-log\frac{exp(\mathbf{c}_k^1\cdot\mathbf{c}_k^2/\tau)}{\sum_{j=1}^Kexp(\mathbf{c}_k^1\cdot\mathbf{c}_j^1/\tau)},
\end{equation}
where $\tau$ is the temperature scaling parameter. In practice, we adopt the symmetric form of Eq. (\ref{loss_clc}) considering the equal importance of the first and the second views.

Finally, weight parameter $\alpha$ is adopted to balance the above two losses:
\begin{equation}
  \label{total_loss}
  \mathcal{L}=\mathcal{L}_{SLA}+\alpha\mathcal{L}_{CLC}.
\end{equation}

In this manner, clustering assignment and embedding optimization are performed alternatively. The overall training detail is shown in Algorithm \ref{spgcc_alg}.

\subsection{Computational Complexity Analysis}
There are several learning stages in SPGCC. At pixel-level pre-training, it takes $\mathcal{O}(\delta^3\gamma^3)$ and $\mathcal{O}(\delta^2\gamma^2)$ for each layer of 3-D and 2-D CNNs, where $\delta$ and $\gamma$ denote the length of feature maps and kernels respectively. It takes $\mathcal{O}(NlogN)$ to perform ERS segmentation on HSI, where $N$ is the number of pixels. For GCN, let $\Vert \mathbf{A} \Vert_0$ be the number of non-zero elements in the adjacency matrix, $L$ be the number of layers, $d$ be the node feature dimension and assumed to be fixed. It takes $\mathcal{O}(L\Vert \mathbf{A} \Vert_0d+LMd^2)$ to aggregate superpixels in the neighbourhood, where $M$ is the number of superpixels. Obtaining clustering results with K-means takes $\mathcal{O}(TKMd)$, where $K$ and $T$ are the number of classes and iterations.

\section{Results and Analysis\label{result}}
\subsection{Datasets}
We evaluate our method on three public HSI datasets, including \textit{India Pines}, \textit{Salinas}, and \textit{Pavia University}. An overview of three datasets can be found in Table \ref{dataset_description}.

\begin{table}[h]
  \caption{Overview of \textit{India Pines}, \textit{Salinas} and \textit{Pavia University}.\label{dataset_description}}
  \centering
  \begin{tblr}{
    row{1} = {c},
    column{2,3,4,5} = {c},
    vline{2} = {-}{0.03em},
    hline{1,5} = {-}{0.08em},
        hline{2} = {-}{0.03em},
      }
    Dataset                   & Size       & \#Band & \#Class & \#Labeled pixels \\
    \textit{India Pines}      & (145, 145) & 200    & 16      & 10249            \\
    \textit{Salinas}          & (512, 217) & 204    & 16      & 54129            \\
    \textit{Pavia University} & (610, 340) & 103    & 9       & 42776
  \end{tblr}
\end{table}

\textit{1) India Pines:} This dataset was captured by the AVIRIS sensor over North-western Indiana in 1992. The size of this scene is 145 $\times$ 145 of which 10249 pixels are labeled. After removing 24 water absorption bands ([104-108], [150-163], 220), each pixel is described by 200 remaining spectral bands. This dataset contains 16 different land cover types.

\textit{2) Salinas:} This dataset was captured by the AVIRIS sensor over Salinas Valley, California. The resolution of this scene is 512 $\times$ 217 with 54129 labeled pixels. Similarly to \textit{India Pines}, 20 water absorption bands ([108-112], [154-167], 224) are removed. \textit{Salinas} contains 204 spectral bands and 16 different land cover types.

\textit{3) Pavia University:} This dataset was collected by the ROSIS sensor over Pavia, northern Italy. It contains 103 spectral bands and 610 $\times$ 340 pixels, of which 42776 are labeled. In this scene, 9 different types of land cover are recorded.

\subsection{Experiment Setup}
\textit{1) Implementation Details:} Our method consists of pixel-level pre-training and superpixel-level contrastive clustering. For the first stage, we divide 3-D pixel cubes with a fixed window size of 27 $\times$ 27 and reduce the spectral dimension to 30 on \textit{India Pines}, 15 on \textit{Salinas}, and 15 on \textit{Pavia University} for computational efficiency. The architecture of CNNs is the same as shown in Table \ref{vae_structure} for three datasets. Adam optimizer is utilized. We set the learning rate and weight decay to $\num{1e-3}$ and $\num{5e-4}$ respectively. When the pre-training of the VAE module is finished, all the pixel cubes are mapped to feature vectors and saved for the following learning stage. For the second stage, we use GCN as the graph encoder, where the sizes of the hidden layer and the output layer are 1024 and 512 respectively. We perform K-means on superpixel embeddings to obtain clustering assignments per five epochs. The temperature scaling parameter of contrast is fixed to 0.5. The weight of clustering-center-level contrast loss is fixed to 0.1. We tune the learning rate $\eta$ from $\num{1e-5}$ to $\num{1e-1}$. Other hyperparameters are shown in Table \ref{hyperparameters}. All the experiments are repeated 10 times and we report the average value and standard deviation of clustering metrics finally.

\begin{table}[ht]
  \caption{Hyperparameter Settings.\label{hyperparameters}}
  \centering
  \begin{tblr}{
    row{1} = {c},
    column{2,3,4,5} = {c},
    vline{2} = {-}{0.03em},
    hline{1,5} = {-}{0.08em},
        hline{2} = {-}{0.03em},
      }
    Dataset                   & $M$  & $\eta$       & $\alpha$ & $\lambda$ & $L$ \\
    \textit{India Pines}      & 1100 & $\num{1e-5}$ & 0.1      & 0.75      & 3   \\
    \textit{Salinas}          & 2700 & $\num{1e-5}$ & 0.1      & 0.55      & 3   \\
    \textit{Pavia University} & 2200 & $\num{1e-4}$ & 0.1      & 0.25      & 3
  \end{tblr}
\end{table}

\textit{2) Compared Methods:} To comprehensively evaluate the performance and applicability of the proposed SPGCC, we introduce ten baselines for comparison, including K-means\cite{kmeans}, fuzzy c-means (FCM)\cite{fuzzycmeans}, sparse subspace clustering (SSC)\cite{ssc}, scalable sparse subspace clustering (SSSC)\cite{sssc}, selective sampling-based scalable sparse subspace clustering (S$^5$C)\cite{s5c}, superpixelwise PCA approach (SuperPCA)\cite{superpca}, graph convolutional optimal transport (GCOT)\cite{gcot}, neighborhood contrastive subspace clustering (NCSC)\cite{ncsc}, superpixel-level global and local similarity graph-based clustering (SGLSC)\cite{sglsc}, dual graph autoencoder (DGAE)\cite{dgae}, spectral–spatial and superpixelwise unsupervised linear discriminant analysis (S$^3$-ULDA)\cite{s3ulda}. Specifically, for K-means and FCM, we perform clustering at the pixel level. For SSC, SSSC, S$^5$C, clustering is performed at the superpixel level to overcome the high memory cost of the subspace clustering. For other methods, we follow the original settings proposed in their papers. A brief summarization of the evaluated methods can be found in Table \ref{method_cls}.

\begin{table}[ht]
  \caption{Properties of the proposed SPGCC and Compared Methods.\label{method_cls}}
  \centering
  \resizebox{\linewidth}{!}{
    \begin{tblr}{
      row{1}={c},
      column{2,3,4,5,6} = {c},
      vline{2} = {-}{0.03em},
      hline{1,14} = {-}{0.08em},
      hline{2,13} = {-}{0.03em},
        }
      Method     & Superpixel   & Subspace clustering & Graph learning & Deep learning & Dimension reduction \\
      K-means    &              &                     &                &                                     \\
      FCM        &              &                     &                &                                     \\
      SSC        & $\checkmark$ & $\checkmark$        &                &                                     \\
      SuperPCA   & $\checkmark$ &                     &                &               & $\checkmark$        \\
      SSSC       & $\checkmark$ & $\checkmark$        &                &                                     \\
      S$^5$C     & $\checkmark$ & $\checkmark$        &                &                                     \\
      GCOT       &              & $\checkmark$        & $\checkmark$   &                                     \\
      SGLSC      & $\checkmark$ & $\checkmark$        & $\checkmark$   &                                     \\
      DGAE       & $\checkmark$ &                     & $\checkmark$   & $\checkmark$                        \\
      NCSC       & $\checkmark$ & $\checkmark$        &                & $\checkmark$                        \\
      S$^3$-ULDA & $\checkmark$ &                     &                &               & $\checkmark$        \\
      SPGCC      & $\checkmark$ &                     & $\checkmark$   & $\checkmark$
    \end{tblr}}
\end{table}

\textit{3) Clustering Metrics:} To quantitatively evaluate the clustering performance, nine popular metrics are utilized: overall accuracy (OA), average accuracy (AA), kappa coefficient (Kappa), normalized mutual information (NMI), adjusted Rand index (ARI), F1-score (F1), Precision, Recall and Purity. Among these metrics, OA, AA, NMI, F1, Precision, Recall, and Purity range in [0\%,100\%] while Kappa and ARI range in [-100\%, 100\%]. For each metric, a higher score indicates a better clustering result. Similar to the common process of clustering evaluation, the Hungarian algorithm \cite{hungarian} is adopted to map the predicted labels according to the ground truth before calculating metrics. In addition, we log the time cost of each method.

\begin{table*}[ht]
  \centering
  \caption{Clustering Performance on HSI Datasets, \textit{India Pines}, \textit{Salinas} and \textit{Pavia University} are Abbreviated to \textit{IP}, \textit{SA} and \textit{PU} Respectively, ``OOM" Denotes ``Out of Memory", and $\bullet$/$\circ$ Indicates whether SPGCC is Statistically Superior/Inferior to the Compared Methods According to the Pairwise $t$-Test at 0.05 Significance Level.\label{clustering_res}}
  \resizebox{\textwidth}{!}{
    \begin{tblr}{
      cells = {c},
      cell{2}{1} = {r=10}{},
      cell{12}{1} = {r=10}{},
      cell{22}{1} = {r=10}{},
      vline{2,3,14} = {0.03em},
      hline{1,32} = {-}{0.08em},
      hline{2,12,22} = {-}{0.03em},
      hline{11,21,31} = {2-14}{dashed},
        }
      Dataset     & Metric (\%) & K-means             & FCM                 & SSC                 & SuperPCA             & SSSC                & S$^5$C              & GCOT                & SGLSC                       & DGAE                & NCSC                & S$^3$-ULDA                  & SPGCC               \\
      \textit{IP} & OA          & 37.37±0.50$\bullet$ & 31.59±0.35$\bullet$ & 47.69±0.60$\bullet$ & 53.21±3.49$\bullet$  & 51.30±2.38$\bullet$ & 44.29±2.70$\bullet$ & 50.30±0.10$\bullet$ & 58.59±1.62$\bullet$         & 58.79±3.12$\bullet$ & 53.42±1.85$\bullet$ & \uline{65.10±2.90}$\bullet$ & \textbf{67.59±1.36} \\
                  & AA          & 41.52±1.67$\bullet$ & 35.41±0.94$\bullet$ & 46.20±4.13$\bullet$ & 50.99±7.24$\bullet$  & 54.83±3.91$\bullet$ & 48.76±3.00$\bullet$ & 37.38±0.31$\bullet$ & \uline{57.03±2.54}$\bullet$ & 52.18±4.71$\bullet$ & 43.82±2.94$\bullet$ & 52.30±3.11$\bullet$         & \textbf{62.74±3.46} \\
                  & Kappa       & 30.45±0.45$\bullet$ & 25.86±0.35$\bullet$ & 41.73±0.79$\bullet$ & 48.58±3.60$\bullet$  & 45.51±2.64$\bullet$ & 38.17±3.20$\bullet$ & 42.45±0.11$\bullet$ & 54.86±1.70$\bullet$         & 54.03±3.36$\bullet$ & 47.69±1.95$\bullet$ & \uline{60.77±2.95}$\bullet$ & \textbf{63.85±1.51} \\
                  & NMI         & 43.94±0.34$\bullet$ & 40.38±0.13$\bullet$ & 57.38±0.71$\bullet$ & 66.39±1.39$\bullet$  & 59.18±1.46$\bullet$ & 55.58±1.14$\bullet$ & 51.77±0.07$\bullet$ & 63.95±0.65$\bullet$         & 64.41±1.13$\bullet$ & 56.75±1.04$\bullet$ & \uline{67.60±1.31}          & \textbf{67.67±1.16} \\
                  & ARI         & 21.86±0.18$\bullet$ & 18.09±0.13$\bullet$ & 31.36±0.63$\bullet$ & 40.61±3.87$\bullet$  & 32.70±1.63$\bullet$ & 30.14±2.15$\bullet$ & 33.22±0.04$\bullet$ & 40.73±0.91$\bullet$         & 47.12±4.03$\bullet$ & 39.10±1.01$\bullet$ & \uline{49.03±3.08}$\bullet$ & \textbf{56.54±1.74} \\
                  & F1          & 30.68±0.39$\bullet$ & 25.94±0.12$\bullet$ & 39.23±0.64$\bullet$ & 47.04±3.43$\bullet$  & 40.52±1.26$\bullet$ & 38.10±1.87$\bullet$ & 42.91±0.03$\bullet$ & 46.55±0.79$\bullet$         & 53.03±3.67$\bullet$ & 46.24±0.93$\bullet$ & \uline{55.03±2.96}$\bullet$ & \textbf{61.26±1.50} \\
                  & Precision   & 33.40±0.55$\bullet$ & 33.22±0.13$\bullet$ & 42.00±0.82$\bullet$ & 53.51±4.88$\bullet$  & 42.88±2.65$\bullet$ & 41.08±2.47$\bullet$ & 36.34±0.04$\bullet$ & \uline{58.19±1.68}$\bullet$ & 58.15±3.62$\bullet$ & 48.42±0.99$\bullet$ & 57.34±2.57$\bullet$         & \textbf{69.30±2.54} \\
                  & Recall      & 28.40±1.06$\bullet$ & 21.28±0.11$\bullet$ & 36.83±1.18$\bullet$ & 42.18±3.74$\bullet$  & 38.50±1.22$\bullet$ & 35.58±1.94$\bullet$ & 52.38±0.03$\bullet$ & 38.81±0.77$\bullet$         & 48.79±3.89$\bullet$ & 44.26±1.28$\bullet$ & \uline{53.15±4.83}$\bullet$ & \textbf{54.91±1.07} \\
                  & Purity      & 51.17±0.21$\bullet$ & 49.95±0.33$\bullet$ & 57.05±1.19$\bullet$ & 68.62±2.69$\bullet$  & 60.31±3.36$\bullet$ & 56.44±2.01$\bullet$ & 54.71±0.09$\bullet$ & 69.95±1.33$\bullet$         & 70.23±1.94$\bullet$ & 61.09±1.75$\bullet$ & \uline{72.32±1.76}$\bullet$ & \textbf{74.27±1.70} \\
                  & Time (s)    & 22.6                & 31.3                & 20.3                & 6.4                  & 2.0                 & 1.4                 & 146.1               & 108.0                       & 25.5                & 77.5                & 11.5                        & 31.1                \\
      \textit{SA} & OA          & 66.97±0.13$\bullet$ & 57.52±3.12$\bullet$ & 71.88±0.92$\bullet$ & 72.58±6.06$\bullet$  & 75.87±1.70$\bullet$ & 61.20±1.71$\bullet$ & OOM                 & \uline{82.09±2.28}$\bullet$ & 66.83±2.67$\bullet$ & 76.47±2.54$\bullet$ & 79.68±2.13$\bullet$         & \textbf{84.47±1.02} \\
                  & AA          & 68.56±0.55$\bullet$ & 63.16±2.70$\bullet$ & 57.94±1.13$\bullet$ & 66.39±5.53$\bullet$  & 63.30±2.52$\bullet$ & 61.03±2.89$\bullet$ & OOM                 & 71.25±0.71                  & 65.77±4.30$\bullet$ & 61.09±0.93$\bullet$ & \uline{71.64±3.20}          & \textbf{72.16±1.46} \\
                  & Kappa       & 63.28±0.10$\bullet$ & 53.63±3.16$\bullet$ & 68.76±1.01$\bullet$ & 69.36±6.92$\bullet$  & 73.04±1.91$\bullet$ & 57.65±1.86$\bullet$ & OOM                 & \uline{80.11±2.41}$\bullet$ & 63.23±2.83$\bullet$ & 73.74±2.72$\bullet$ & 77.18±2.52$\bullet$         & \textbf{82.69±1.13} \\
                  & NMI         & 72.45±0.15$\bullet$ & 69.35±0.99$\bullet$ & 80.51±0.65$\bullet$ & 82.67±1.43$\bullet$  & 83.44±1.18$\bullet$ & 76.23±0.97$\bullet$ & OOM                 & \textbf{87.57±0.58}$\circ$  & 78.03±1.51$\bullet$ & 83.61±2.05$\bullet$ & \uline{87.41±2.13}          & 86.88±0.56          \\
                  & ARI         & 52.55±0.18$\bullet$ & 45.07±2.53$\bullet$ & 68.82±0.73$\bullet$ & 66.39±6.47$\bullet$  & 68.00±1.54$\bullet$ & 50.14±1.35$\bullet$ & OOM                 & \uline{71.64±2.27}$\bullet$ & 56.18±3.10$\bullet$ & 69.02±5.61$\bullet$ & 67.00±3.44$\bullet$         & \textbf{75.98±1.46} \\
                  & F1          & 57.62±0.20$\bullet$ & 50.17±2.49$\bullet$ & 72.14±0.64$\bullet$ & 70.30±5.43$\bullet$  & 71.49±1.35$\bullet$ & 54.91±1.18$\bullet$ & OOM                 & \uline{74.58±2.08}$\bullet$ & 60.69±3.14$\bullet$ & 72.40±5.07$\bullet$ & 70.79±3.12$\bullet$         & \textbf{78.50±1.30} \\
                  & Precision   & 54.98±0.31$\bullet$ & 55.00±1.57$\bullet$ & 69.03±0.90$\bullet$ & 63.11±10.43$\bullet$ & 66.83±1.70$\bullet$ & 58.52±1.83$\bullet$ & OOM                 & \uline{73.30±2.39}$\bullet$ & 59.98±2.33$\bullet$ & 67.19±3.44$\bullet$ & 63.04±5.70$\bullet$         & \textbf{76.33±1.36} \\
                  & Recall      & 60.53±0.86$\bullet$ & 46.28±4.01$\bullet$ & 75.55±0.71$\bullet$ & \uline{80.85±2.49}   & 76.89±1.55$\bullet$ & 51.74±0.91$\bullet$ & OOM                 & 76.04±3.74$\bullet$         & 62.27±8.68$\bullet$ & 78.60±7.42          & \textbf{82.75±2.39}$\circ$  & 80.81±1.53          \\
                  & Purity      & 67.36±0.18$\bullet$ & 67.07±0.96$\bullet$ & 76.13±1.42$\bullet$ & 74.30±5.23$\bullet$  & 77.15±1.52$\bullet$ & 70.68±1.66$\bullet$ & OOM                 & \uline{83.57±1.34}          & 72.16±1.36$\bullet$ & 77.79±2.40$\bullet$ & 79.76±2.08$\bullet$         & \textbf{84.55±0.95} \\
                  & Time (s)    & 109.5               & 155.2               & 11.7                & 92.1                 & 12.4                & 2.4                 & OOM                 & 47.0                        & 360.2               & 2408.3              & 194.0                       & 150.6               \\
      \textit{PU} & OA          & 53.41±0.00$\bullet$ & 51.62±1.00$\bullet$ & 55.17±0.00$\bullet$ & 46.66±1.77$\bullet$  & 48.77±0.66$\bullet$ & 46.01±0.58$\bullet$ & OOM                 & 57.18±0.29$\bullet$         & 59.53±4.08$\bullet$ & OOM                 & \uline{59.94±0.59}$\bullet$ & \textbf{62.68±0.48} \\
                  & AA          & 52.59±0.01$\bullet$ & 55.31±0.85$\bullet$ & 48.81±0.00$\bullet$ & 52.05±2.28$\bullet$  & 50.73±2.27$\bullet$ & 46.06±0.84$\bullet$ & OOM                 & 53.67±0.29$\bullet$         & 51.80±3.03$\bullet$ & OOM                 & \uline{56.34±0.83}$\bullet$ & \textbf{59.03±0.53} \\
                  & Kappa       & 43.36±0.00$\bullet$ & 42.09±1.10$\bullet$ & 45.63±0.00$\bullet$ & 36.10±1.87$\bullet$  & 40.02±0.67$\bullet$ & 34.76±0.36$\bullet$ & OOM                 & 48.22±0.31$\bullet$         & 49.94±3.97$\bullet$ & OOM                 & \uline{51.46±0.63}$\bullet$ & \textbf{54.91±0.56} \\
                  & NMI         & 53.30±0.00$\bullet$ & 52.84±0.81$\bullet$ & 51.48±0.00$\bullet$ & 44.24±0.04$\bullet$  & 49.22±0.98$\bullet$ & 47.41±0.28$\bullet$ & OOM                 & 53.89±0.43$\bullet$         & 53.48±2.08$\bullet$ & OOM                 & \textbf{61.15±0.25}$\circ$  & \uline{57.92±1.14}  \\
                  & ARI         & 32.13±0.00$\bullet$ & 30.52±0.49$\bullet$ & 35.26±0.00$\bullet$ & 25.42±1.00$\bullet$  & 29.60±0.63$\bullet$ & 28.97±1.27$\bullet$ & OOM                 & 42.76±0.25$\bullet$         & 44.83±5.15          & OOM                 & \textbf{49.37±0.38}$\circ$  & \uline{46.30±0.91}  \\
                  & F1          & 45.43±0.00$\bullet$ & 43.42±0.65$\bullet$ & 50.04±0.00$\bullet$ & 39.66±0.91$\bullet$  & 42.39±0.45$\bullet$ & 43.36±1.35$\bullet$ & OOM                 & 54.08±0.16$\bullet$         & 56.49±4.76          & OOM                 & \textbf{58.20±0.37}$\circ$  & \uline{56.79±0.70}  \\
                  & Precision   & 56.88±0.00$\bullet$ & 57.37±0.37$\bullet$ & 53.78±0.00$\bullet$ & 50.92±0.79$\bullet$  & 57.26±1.06$\bullet$ & 52.49±0.50$\bullet$ & OOM                 & 67.11±0.42$\bullet$         & 65.24±2.97$\bullet$ & OOM                 & \uline{70.75±0.34}          & \textbf{71.32±1.16} \\
                  & Recall      & 37.82±0.00$\bullet$ & 34.95±0.94$\bullet$ & 46.78±0.00$\bullet$ & 32.49±0.90$\bullet$  & 33.65±0.40$\bullet$ & 36.95±1.66$\bullet$ & OOM                 & 45.28±0.08$\bullet$         & \textbf{49.95±5.57} & OOM                 & \uline{49.44±0.55}$\circ$   & 47.19±0.55          \\
                  & Purity      & 69.61±0.00$\bullet$ & 69.30±0.95$\bullet$ & 63.87±0.00$\bullet$ & 59.44±0.94$\bullet$  & 65.93±0.95$\bullet$ & 66.19±0.00$\bullet$ & OOM                 & 71.90±0.22$\bullet$         & 70.57±1.63$\bullet$ & OOM                 & \uline{76.54±0.39}$\bullet$ & \textbf{77.31±0.75} \\
                  & Time (s)    & 113.7               & 82.2                & 16.4                & 87.4                 & 7.7                 & 2.9                 & OOM                 & 195.6                       & 202.0               & OOM                 & 374.5                       & 103.5
    \end{tblr}}
\end{table*}

\begin{figure*}[ht]
  \centering
  \subfloat[]{\includegraphics[width=0.0765\textwidth]{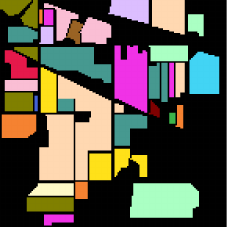}\label{IP_gt}}
  \hfill
  \subfloat[]{\includegraphics[width=0.0765\textwidth]{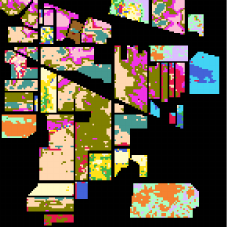}\label{IP_kmeans}}
  \hfill
  \subfloat[]{\includegraphics[width=0.0765\textwidth]{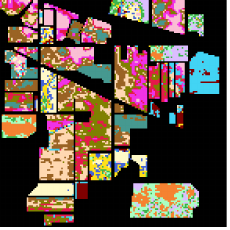}\label{IP_fcm}}
  \hfill
  \subfloat[]{\includegraphics[width=0.0765\textwidth]{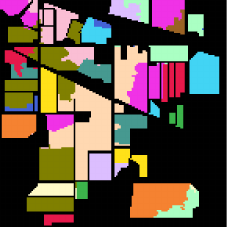}\label{IP_SSC}}
  \hfill
  \subfloat[]{\includegraphics[width=0.0765\textwidth]{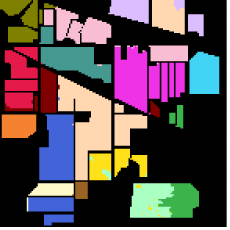}\label{IP_SuperPCA}}
  \hfill
  \subfloat[]{\includegraphics[width=0.0765\textwidth]{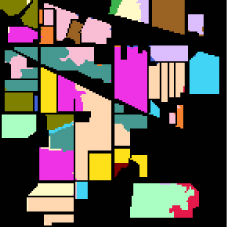}\label{IP_SSSC}}
  \hfill
  \subfloat[]{\includegraphics[width=0.0765\textwidth]{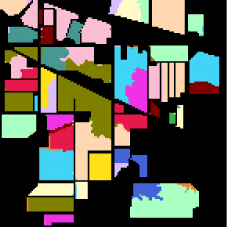}\label{IP_S5C}}
  \hfill
  \subfloat[]{\includegraphics[width=0.0765\textwidth]{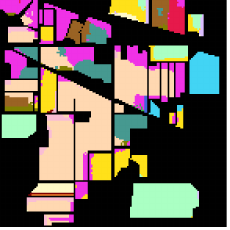}\label{IP_GCOT}}
  \hfill
  \subfloat[]{\includegraphics[width=0.0765\textwidth]{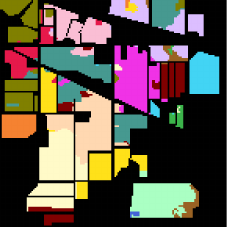}\label{IP_SGLSC}}
  \hfill
  \subfloat[]{\includegraphics[width=0.0765\textwidth]{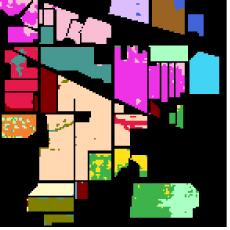}\label{IP_DGAE}}
  \hfill
  \subfloat[]{\includegraphics[width=0.0765\textwidth]{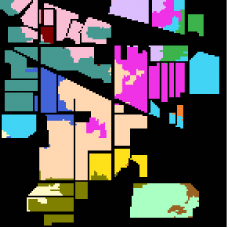}\label{IP_NCSC}}
  \hfill
  \subfloat[]{\includegraphics[width=0.0765\textwidth]{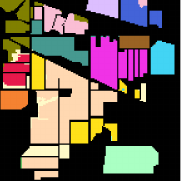}\label{IP_S3ULDA}}
  \hfill
  \subfloat[]{\includegraphics[width=0.0765\textwidth]{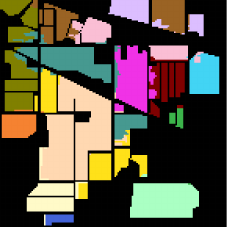}\label{IP_SPGCC}}
  \caption{Clustering maps on \textit{India Pines} (a) ground truth, (b) K-means, (c) FCM, (d) SSC, (e) SuperPCA, (f) SSSC, (g) S$^5$C, (h) GCOT, (i) SGLSC, (j) DGAE, (k) NCSC, (l) S$^3$-ULDA, (m) SPGCC.}
  \label{IP_clustering_map}
\end{figure*}

\begin{figure*}[ht]
  \centering
  \subfloat[]{\includegraphics[width=0.083\textwidth]{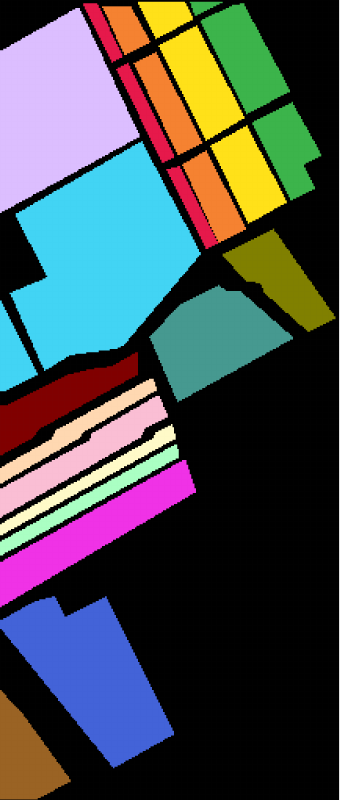}\label{SA_gt}}
  \hfill
  \subfloat[]{\includegraphics[width=0.083\textwidth]{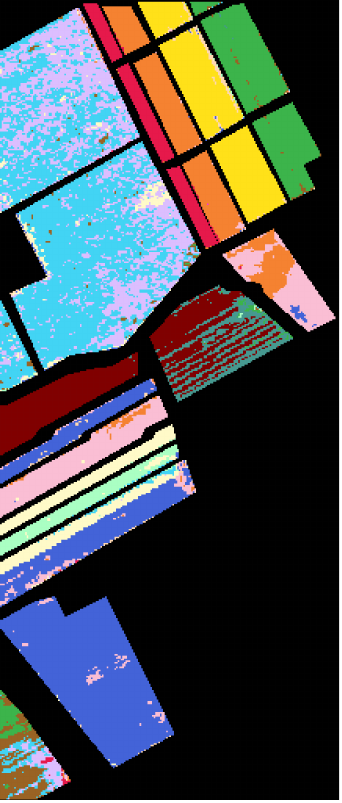}\label{SA_kmeans}}
  \hfill
  \subfloat[]{\includegraphics[width=0.083\textwidth]{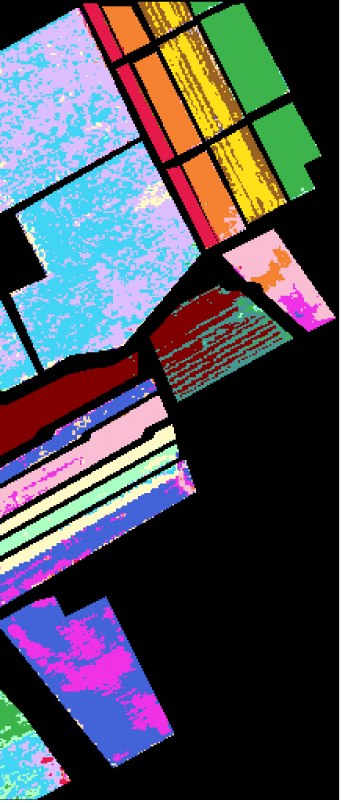}\label{SA_fcm}}
  \hfill
  \subfloat[]{\includegraphics[width=0.083\textwidth]{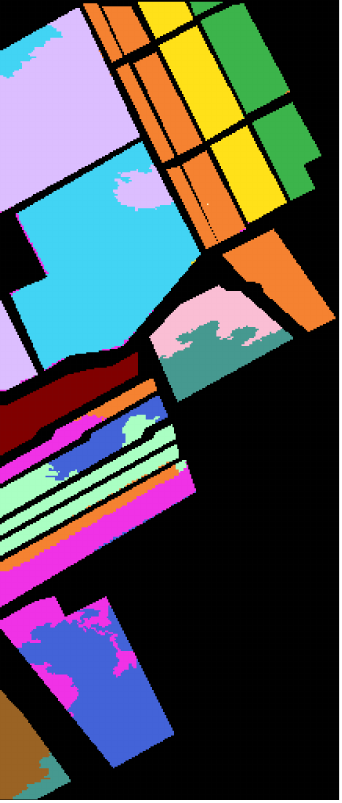}\label{SA_SSC}}
  \hfill
  \subfloat[]{\includegraphics[width=0.083\textwidth]{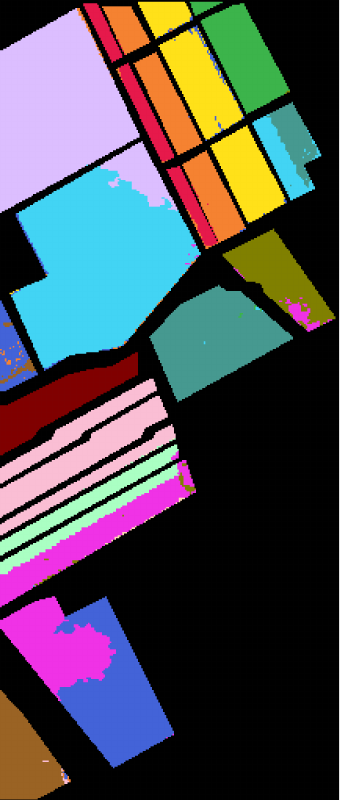}\label{SA_SuperPCA}}
  \hfill
  \subfloat[]{\includegraphics[width=0.083\textwidth]{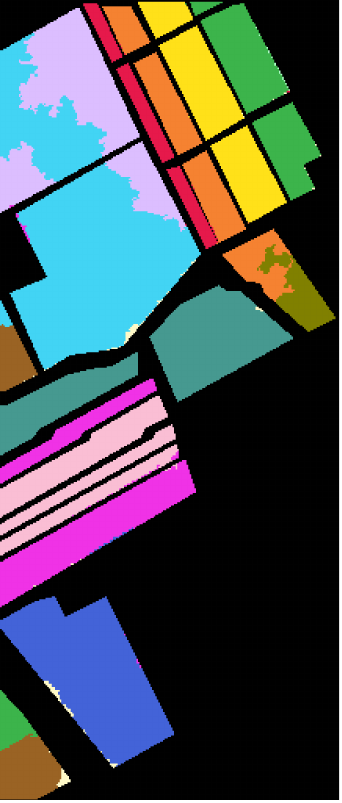}\label{SA_SSSC}}
  \hfill
  \subfloat[]{\includegraphics[width=0.083\textwidth]{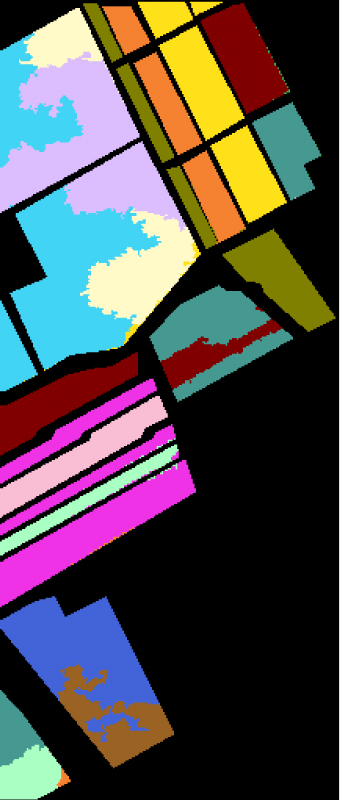}\label{SA_S5C}}
  \hfill
  \subfloat[]{\includegraphics[width=0.083\textwidth]{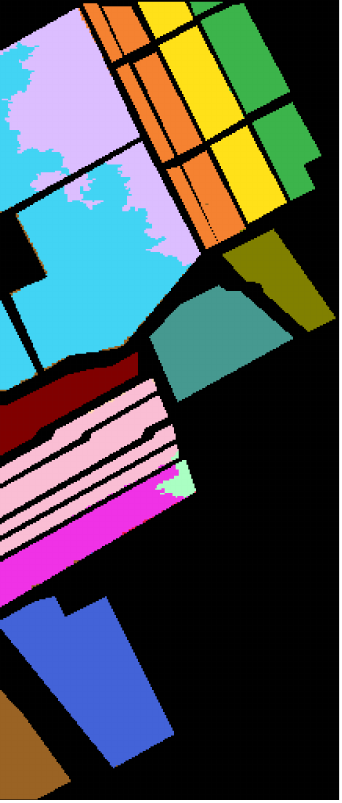}\label{SA_SGLSC}}
  \hfill
  \subfloat[]{\includegraphics[width=0.083\textwidth]{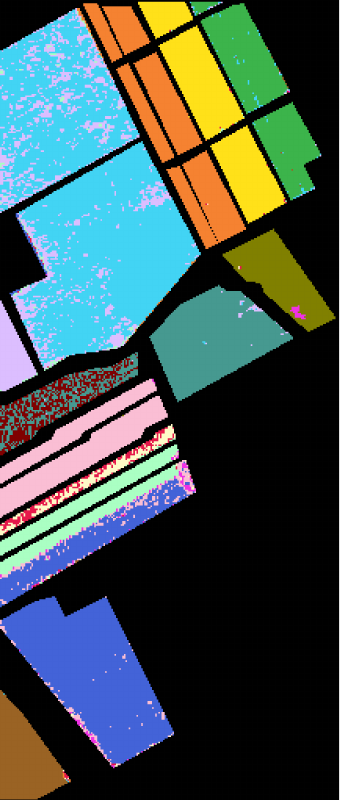}\label{SA_DGAE}}
  \hfill
  \subfloat[]{\includegraphics[width=0.083\textwidth]{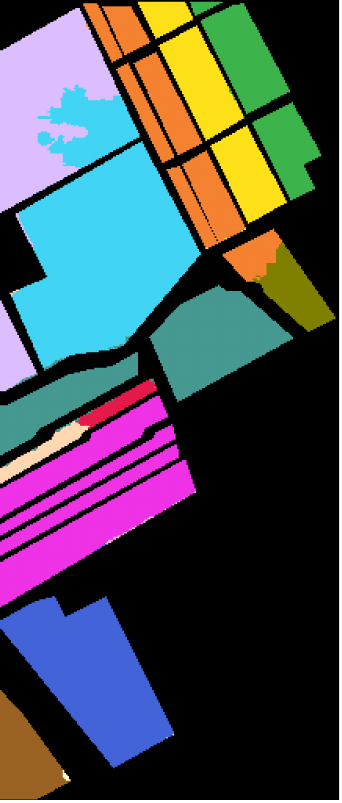}\label{SA_NCSC}}
  \hfill
  \subfloat[]{\includegraphics[width=0.083\textwidth]{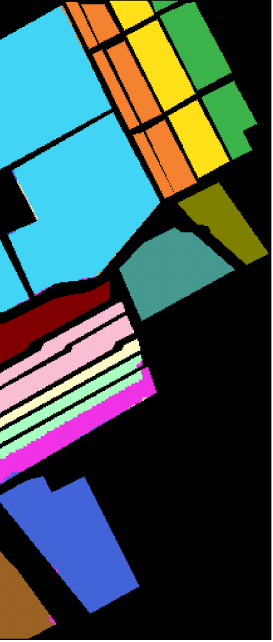}\label{SA_S3ULDA}}
  \hfill
  \subfloat[]{\includegraphics[width=0.083\textwidth]{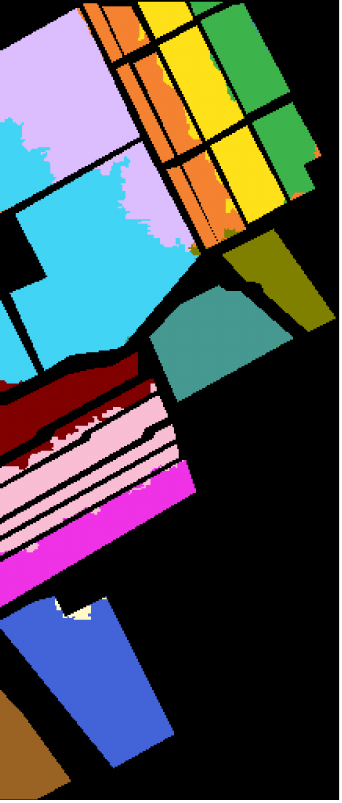}\label{SA_SPGCC}}
  \caption{Clustering maps on \textit{Salinas} (a) ground truth, (b) K-means, (c) FCM, (d) SSC, (e) SuperPCA, (f) SSSC, (g) S$^5$C, (h) SGLSC, (i) DGAE, (j) NCSC, (k) S$^3$-ULDA, (l) SPGCC.}
  \label{SA_clustering_map}
\end{figure*}

\begin{figure*}[ht]
  \centering
  \subfloat[]{\includegraphics[width=0.09\textwidth]{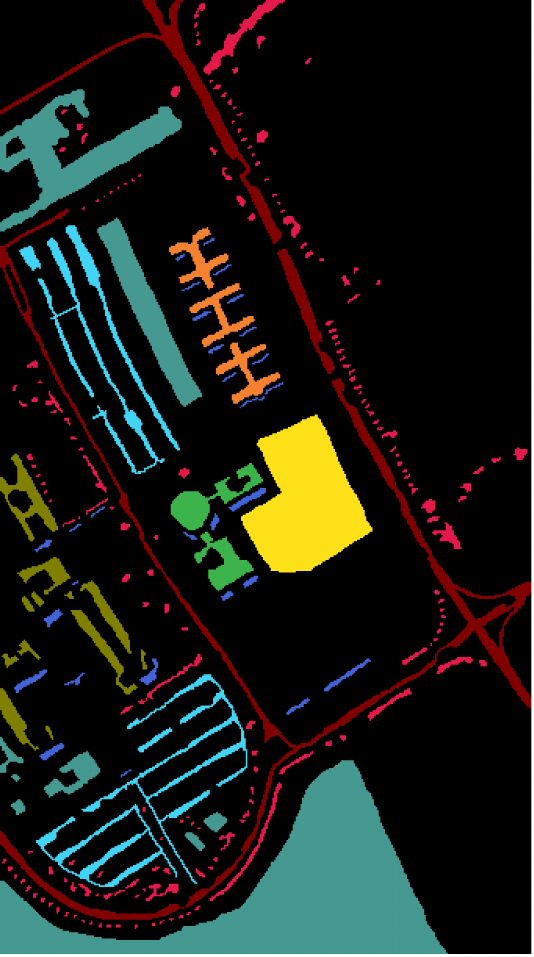}\label{PU_gt}}
  \hfill
  \subfloat[]{\includegraphics[width=0.09\textwidth]{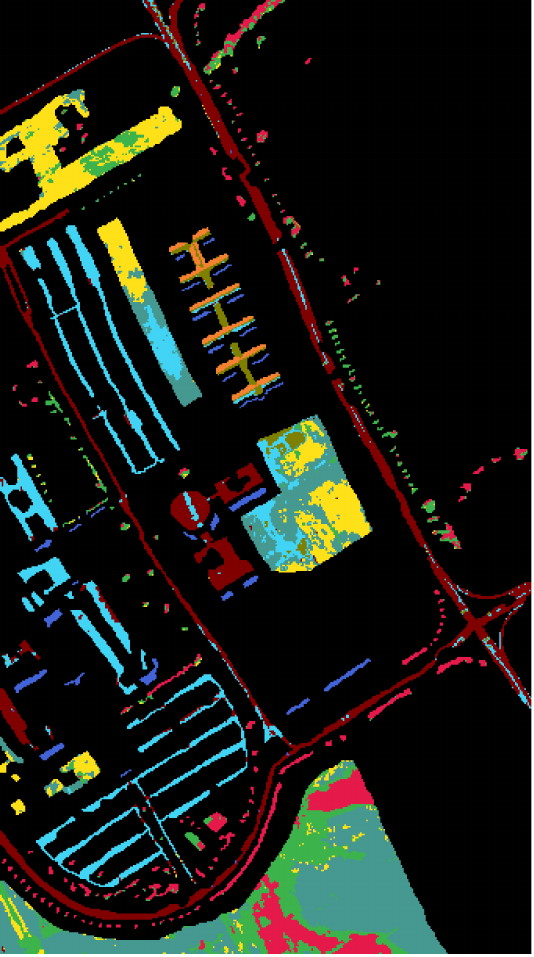}\label{PU_kmeans}}
  \hfill
  \subfloat[]{\includegraphics[width=0.09\textwidth]{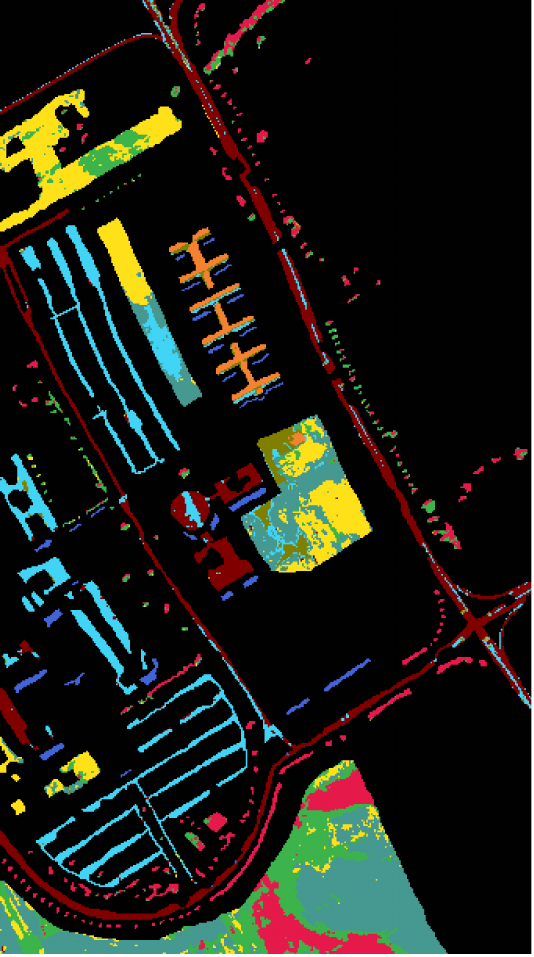}\label{PU_fcm}}
  \hfill
  \subfloat[]{\includegraphics[width=0.09\textwidth]{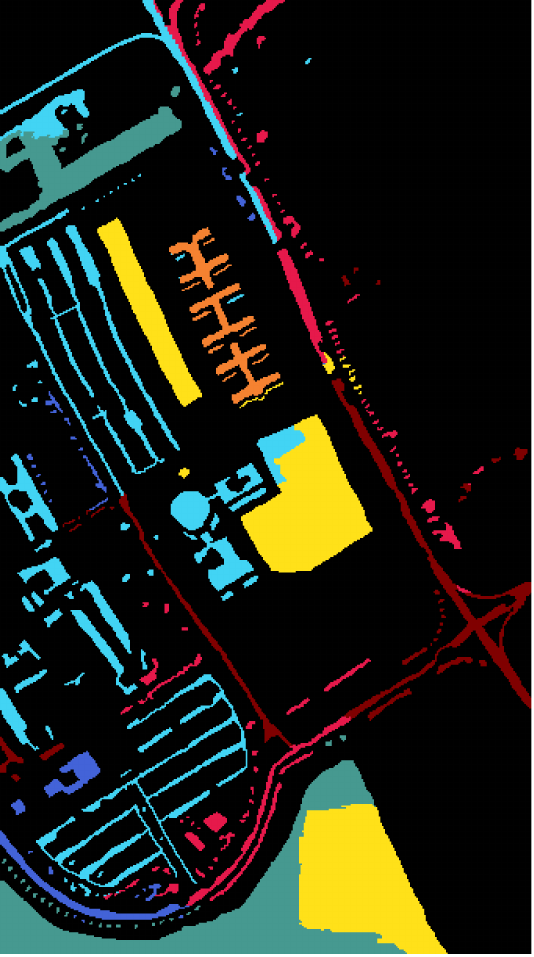}\label{PU_SSC}}
  \hfill
  \subfloat[]{\includegraphics[width=0.09\textwidth]{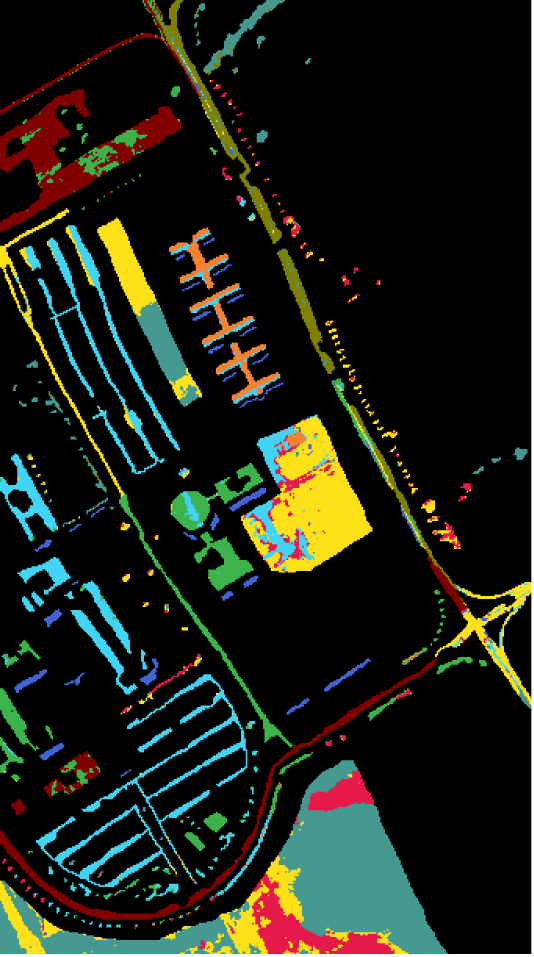}\label{PU_SuperPCA}}
  \hfill
  \subfloat[]{\includegraphics[width=0.09\textwidth]{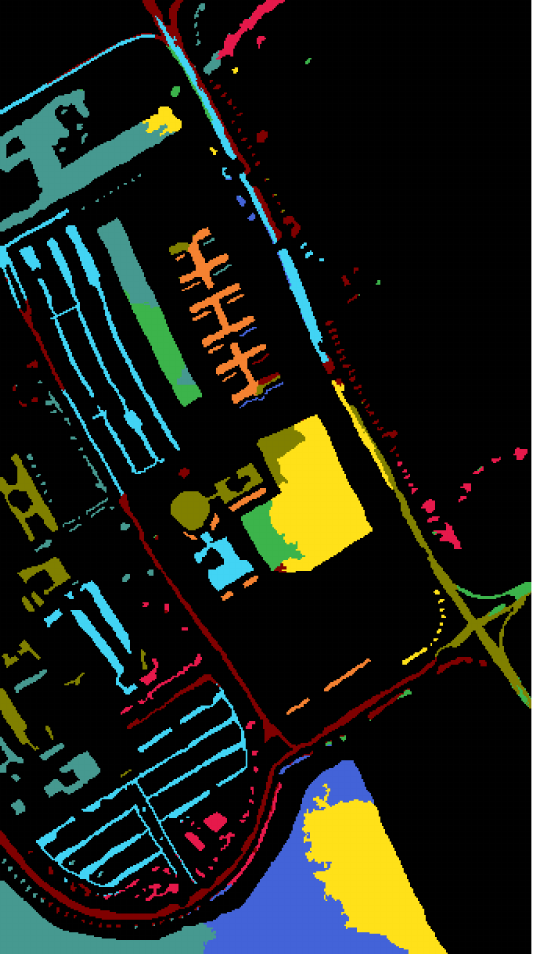}\label{PU_SSSC}}
  \hfill
  \subfloat[]{\includegraphics[width=0.09\textwidth]{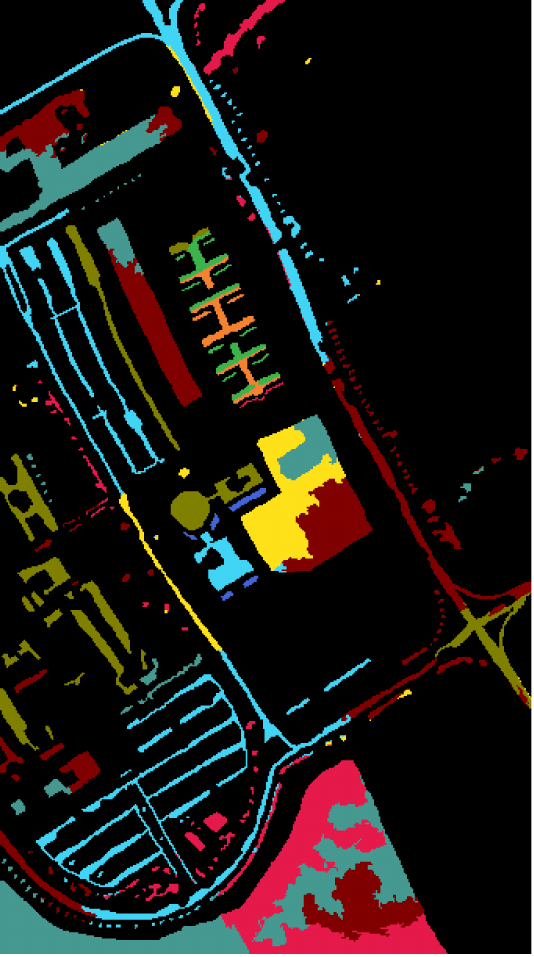}\label{PU_S5C}}
  \hfill
  \subfloat[]{\includegraphics[width=0.09\textwidth]{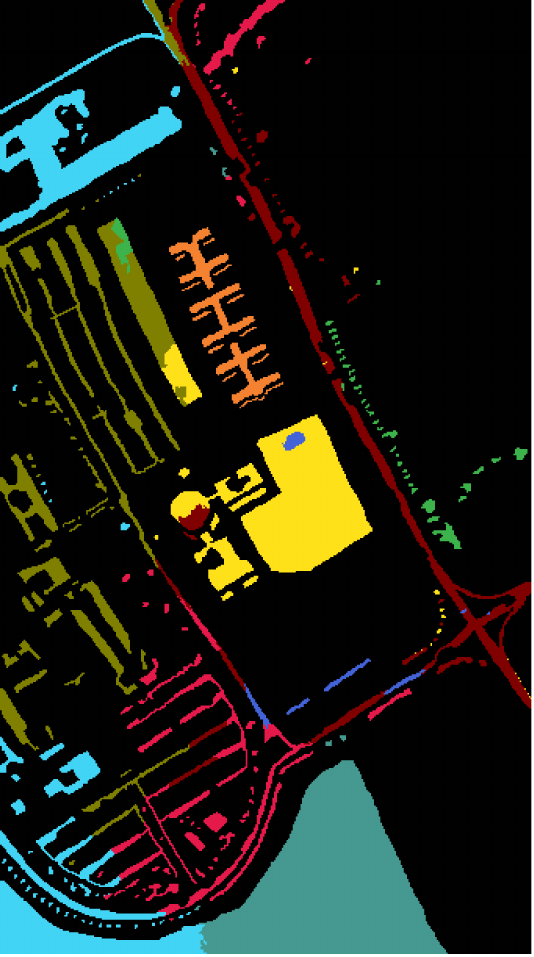}\label{PU_SGLSC}}
  \hfill
  \subfloat[]{\includegraphics[width=0.09\textwidth]{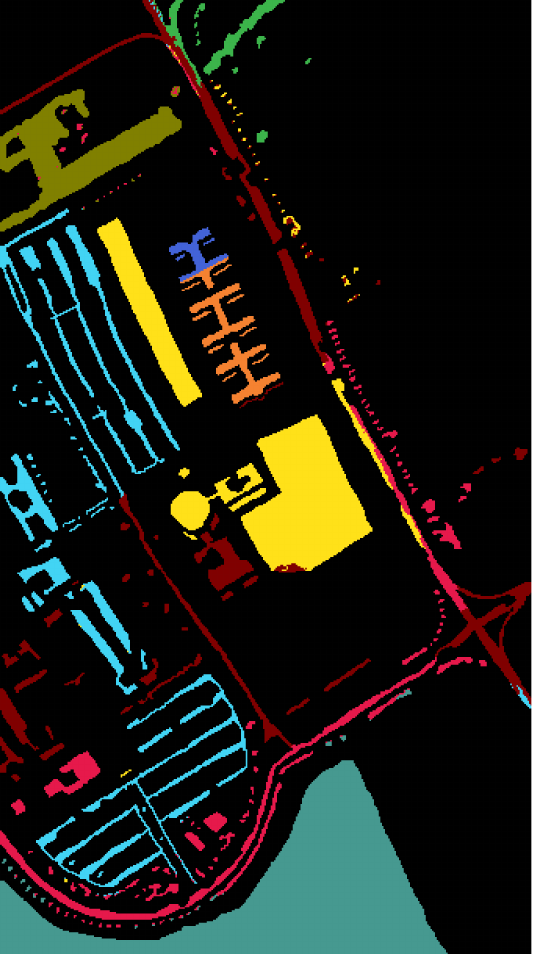}\label{PU_DGAE}}
  \hfill
  \subfloat[]{\includegraphics[width=0.09\textwidth]{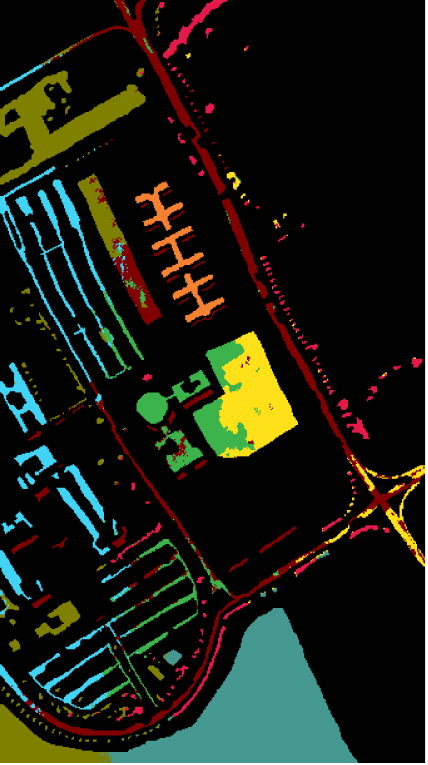}\label{PU_S3ULDA}}
  \hfill
  \subfloat[]{\includegraphics[width=0.09\textwidth]{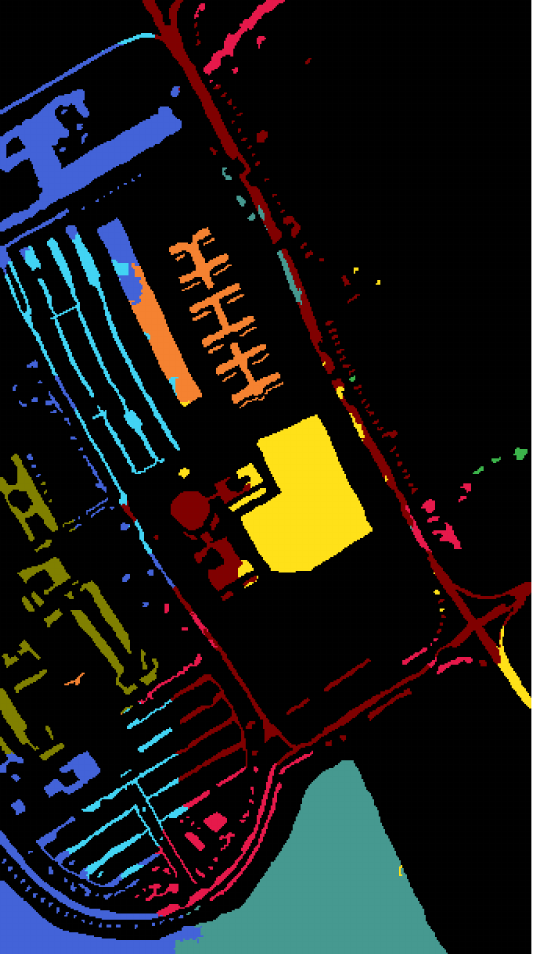}\label{PU_SPGCC}}
  \caption{Clustering maps on \textit{Pavia University} (a) ground truth, (b) K-means, (c) FCM, (d) SSC, (e) SuperPCA, (f) SSSC, (g) S$^5$C, (h) SGLSC, (i) DGAE, (j) S$^3$-ULDA, (k) SPGCC.}
  \label{PU_clustering_map}
\end{figure*}

\begin{figure*}[ht]
  \centering
  \subfloat[]{\includegraphics[width=0.32\textwidth]{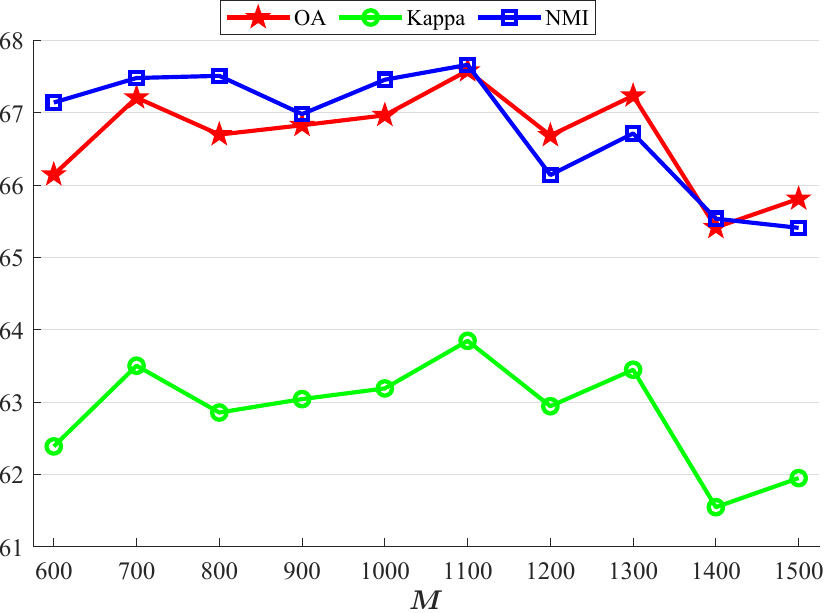}\label{IP_param_n_sp}}
  \hfill
  \subfloat[]{\includegraphics[width=0.32\textwidth]{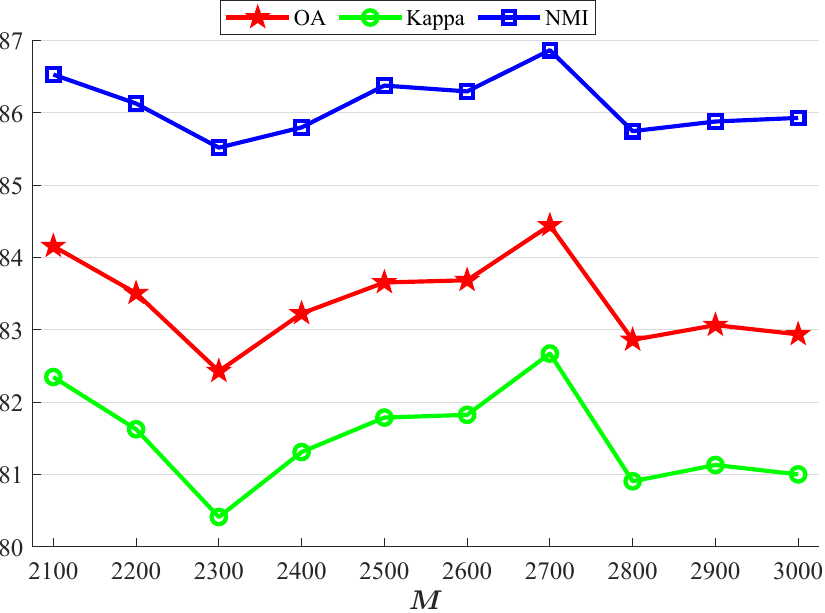}\label{SA_param_n_sp}}
  \hfill
  \subfloat[]{\includegraphics[width=0.32\textwidth]{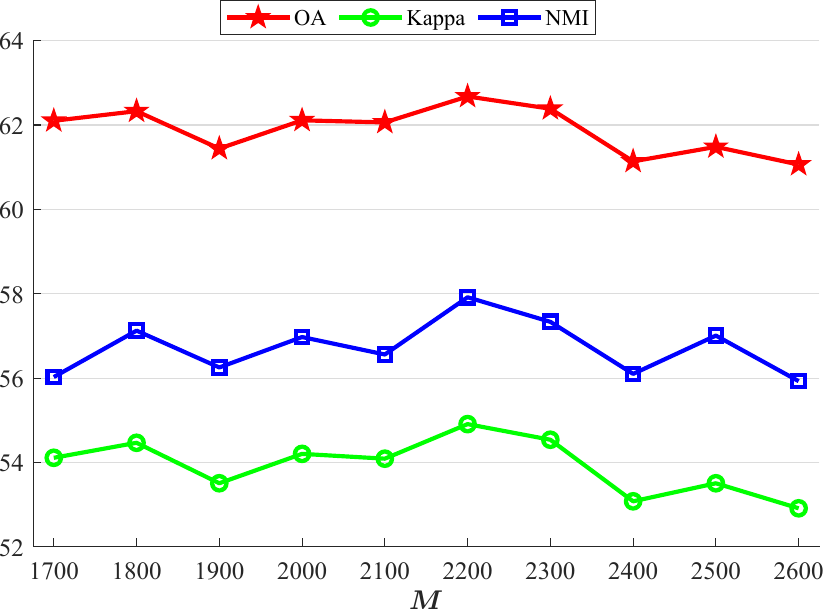}\label{PU_param_n_sp}}
  \caption{The influence of different numbers of superpixels $M$ on (a) \textit{India Pines}, (b) \textit{Salinas}, (c) \textit{Pavia University}.}
  \label{param_n_sp}
\end{figure*}

\begin{figure*}[ht]
  \centering
  \subfloat[]{\includegraphics[width=0.32\textwidth]{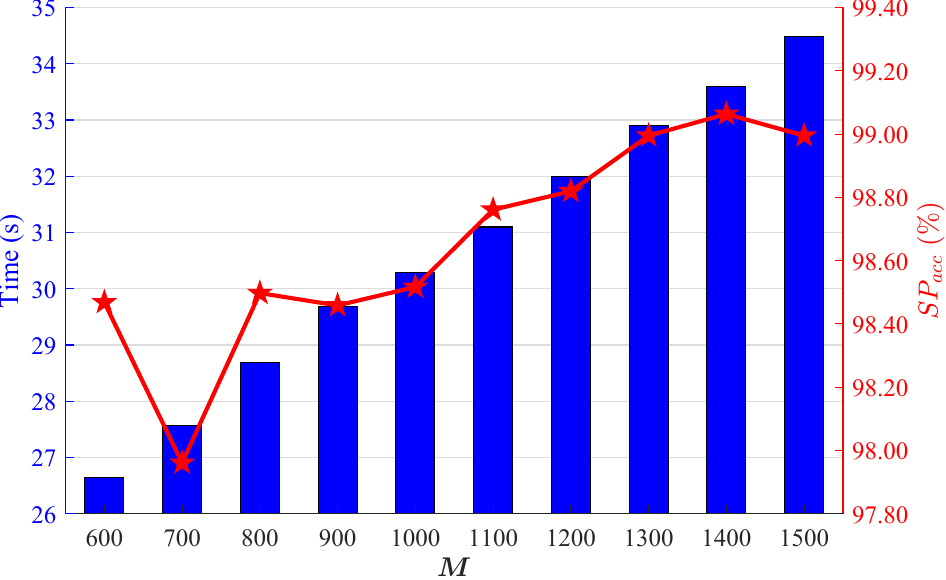}\label{IP_time}}
  \hfill
  \subfloat[]{\includegraphics[width=0.32\textwidth]{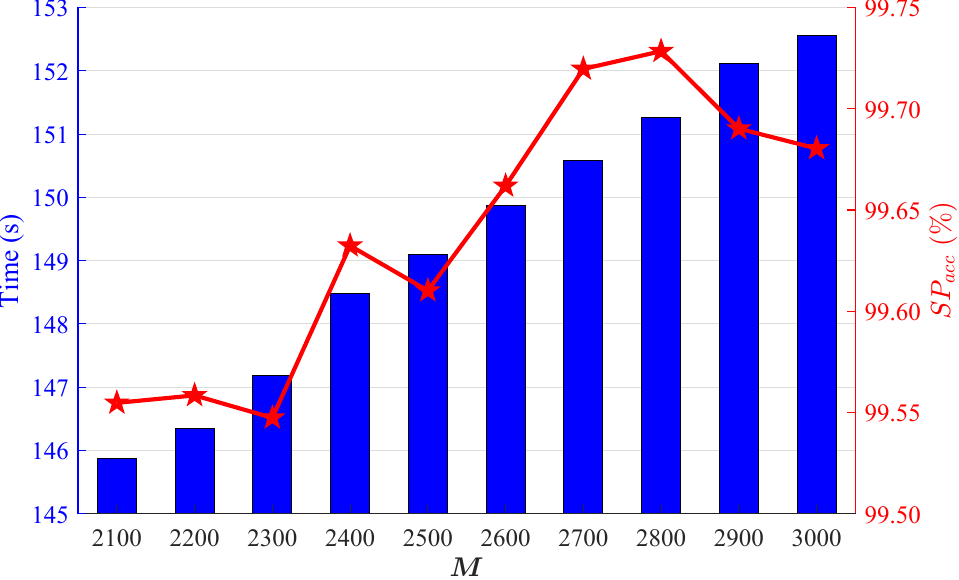}\label{SA_time}}
  \hfill
  \subfloat[]{\includegraphics[width=0.32\textwidth]{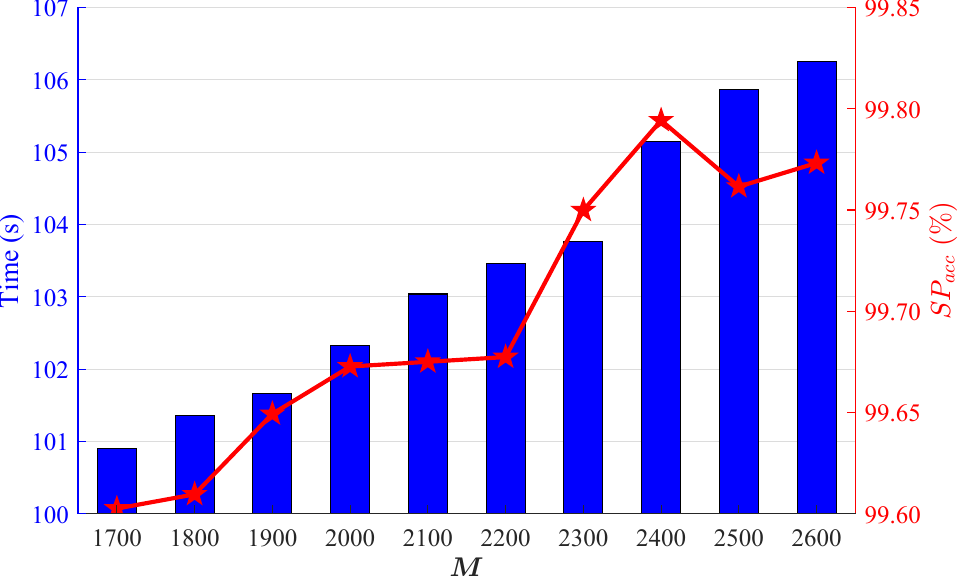}\label{PU_time}}
  \caption{Time cost and superpixel segmentation accuracy for different numbers of superpixels $M$ on (a) \textit{India Pines}, (b) \textit{Salinas}, (c) \textit{Pavia University}.}
  \label{time_sp_acc}
\end{figure*}

\subsection{Comparison with Other Methods}
In Table \ref{clustering_res}, we report the clustering performance and the running time of SPGCC together with compared methods on \textit{India Pines}, \textit{Salinas}, and \textit{Pavia University}. For each clustering metric, the optimal value is marked in bold, and the suboptimal value is underlined. Clustering map visualizations of different methods are shown in Figs. \ref{IP_clustering_map}, \ref{SA_clustering_map}, \ref{PU_clustering_map}. All the experiments were conducted in a Linux server with an AMD EPYC 7352 CPU and an NVIDIA RTX 3090 GPU. CUDA was available for deep learning methods except NCSC on \textit{Salinas} because of the huge cost of memory.

According to the quantitative metrics and clustering maps, it could be easily observed that the proposed SPGCC method outperforms other compared methods on three datasets significantly. For example, OA obtained by SPGCC has an improvement of 2.5\%, 2.4\%, and 2.7\% over the second-best method on the three datasets, respectively. Results of the significance test further validate the superiority of SPGCC (i.e., eight metrics on \textit{India Pines}, five metrics on \textit{Salinas}, and four metrics on \textit{Pavia University} have statistical superiority to the suboptimal values of different compared methods).

GCOT and NCSC fail to run on \textit{Salinas} or \textit{Pavia University} due to the extreme cost of memory. GCOT does not use superpixel segmentation and therefore is not suitable for large-scale HSI datasets, indicating the necessity of using superpixel segmentation. Despite the risk of grouping pixels of different classes into the same superpixel, superpixel segmentation brings high homogeneity, and we have verified that the accuracy of segmentation is generally above 99\% (in Section \ref{sp_num_acc}). NCSC needs to perform convolution on all pixel cubes in a single batch due to the network structure. Specifically, although NCSC only conducts lightweight 2-D CNNs and introduces superpixel segmentation to reduce the scale of subspace clustering, the pixel-level convolutional network and the superpixel-level clustering network are not separated, resulting in high computational costs. In contrast, the proposed SPGCC also performs pixel-level convolution but in the pre-training stage which is separated from the subsequent network. As a result, SPGCC could extract high-order spectral and spatial features at the pixel level and utilize superpixel segmentation to reduce data scale simultaneously.

Compared with the recent graph learning methods (i.e., SGLSC and DGAE), SPGCC performs better on all three datasets, with similar or less time cost. SPGCC uses 3-D and 2-D hybrid CNNs at the pre-training stage to extract high-order spatial and spectral features of HSI pixels, which is conducive to subsequent clustering, while SGLSC performs clustering on raw HSI features. Moreover, SPGCC is based on contrastive clustering and has an advantage over the GAE-based method because of the clustering-oriented optimization target (i.e., pulling superpixel representations of the same class close, while pushing different clusters' representations away). In this manner, SPGCC gains clustering-friendly representations with better intra-class similarity and inter-class dissimilarity.

\subsection{Parameters Analysis\label{sp_num_acc}}
Our method mainly contains four hyperparameters: number of superpixels ($M$), weight of clustering-center-level contrast loss ($\alpha$), ratio of high-confidence samples ($\lambda$) and number of GCN layers ($L$). They have been tuned for different datasets. The adopted values are shown in Table \ref{hyperparameters}.

\textit{1) Impact of $M$:} The number of superpixels $M$ controls the scale and homogeneity of superpixels. The influence of $M$ on clustering performance is shown in Fig. \ref{param_n_sp}. It can be seen that the proposed SPGCC has robustness for different $M$. As $M$ increases, different metrics almost show an increasing-then-decreasing trend. The optimal $M$ are 1100, 2700, and 2200 for three datasets, respectively.

Moreover, we validate the effect of $M$ on the running time and the superpixel segmentation accuracy in Fig. \ref{time_sp_acc}. Specifically, we fix the training epochs to 200 and record the running time. Due to the efficient parallelism of GCN, the increase in time consumption is lower than that of $M$. As for the superpixel segmentation accuracy, we define a metric called average superpixel segmentation accuracy $SP^{Avg}_{acc}$ as:
\begin{equation}
  \label{sp_acc}
  SP^{Avg}_{acc}=\frac{\sum_{i=1}^Mn_{i}^{dominant}}{\sum_{i=1}^Mn_{i}^{labeled}},
\end{equation}
where $n_{i}^{dominant}$ and $n_{i}^{labeled}$ are the numbers of labeled pixels in the dominant class and total labeled pixels within the $i$-th superpixel, respectively. With the increase of $M$, $SP^{Avg}_{acc}$ shows a gradual upward trend, indicating that pixels within one superpixel are more likely to belong to the same class in a higher granularity partition. Since $SP^{Avg}_{acc}$ typically exceeds 99\%, different $M$ have minimal impact on the accuracy of superpixel segmentation.

\begin{figure*}[ht]
  \centering
  \subfloat[]{\includegraphics[width=0.32\textwidth]{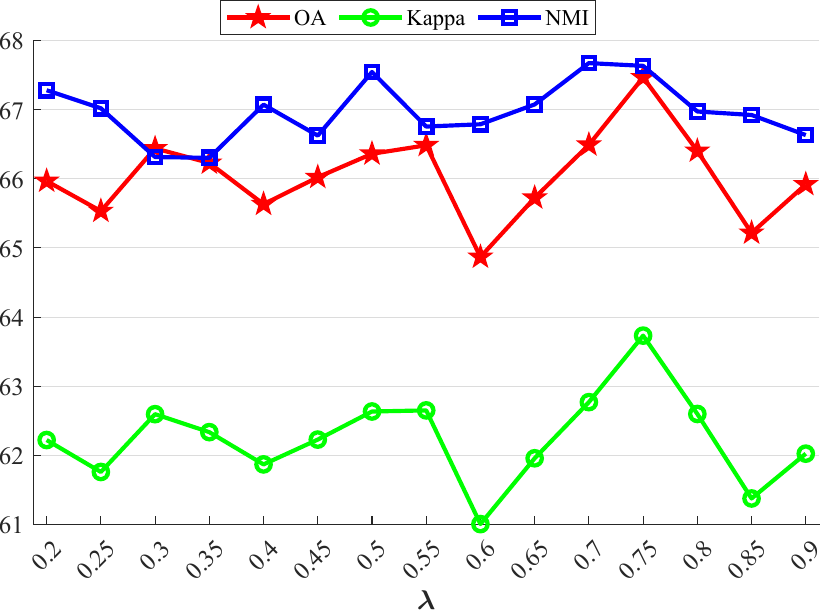}\label{IP_param_lambda}}
  \hfill
  \subfloat[]{\includegraphics[width=0.32\textwidth]{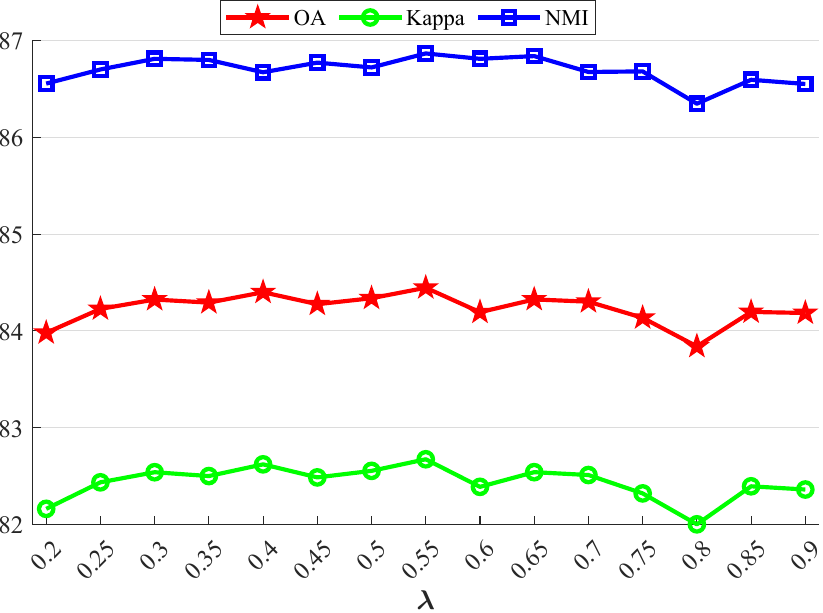}\label{SA_param_lambda}}
  \hfill
  \subfloat[]{\includegraphics[width=0.32\textwidth]{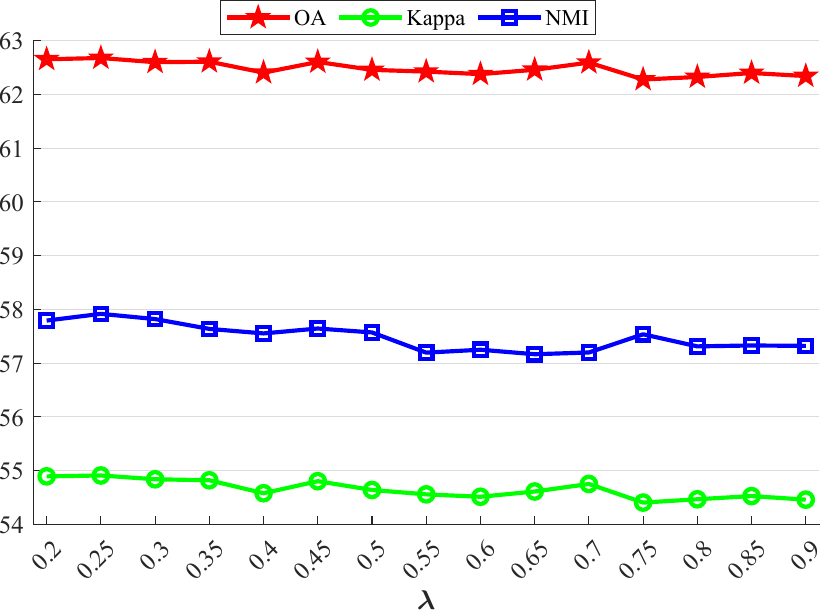}\label{PU_param_lambda}}
  \caption{The influence of different ratio of selected high-confidence samples $\lambda$ on (a) \textit{India Pines}, (b) \textit{Salinas}, (c) \textit{Pavia University}.}
  \label{param_lambda}
\end{figure*}

\begin{figure*}[ht]
  \centering
  \subfloat[]{\includegraphics[width=0.32\textwidth]{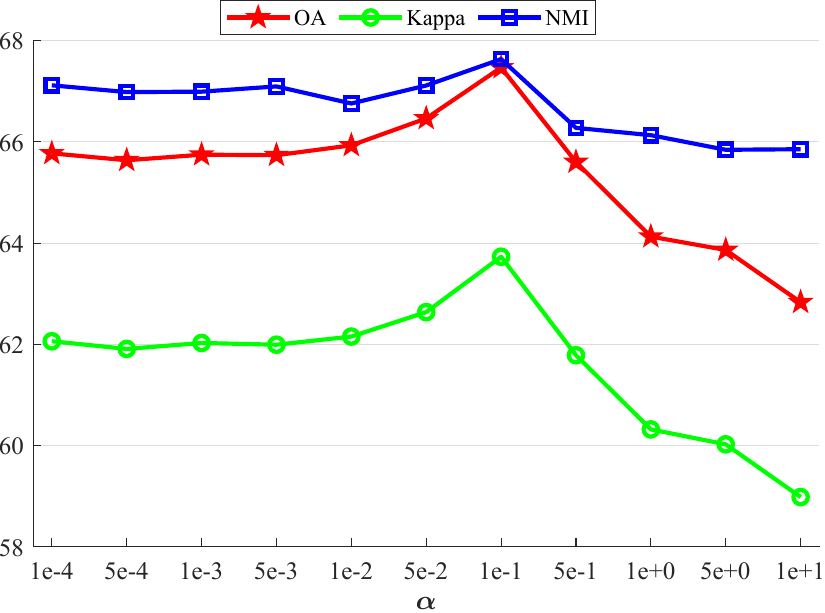}\label{IP_param_alpha}}
  \hfill
  \subfloat[]{\includegraphics[width=0.32\textwidth]{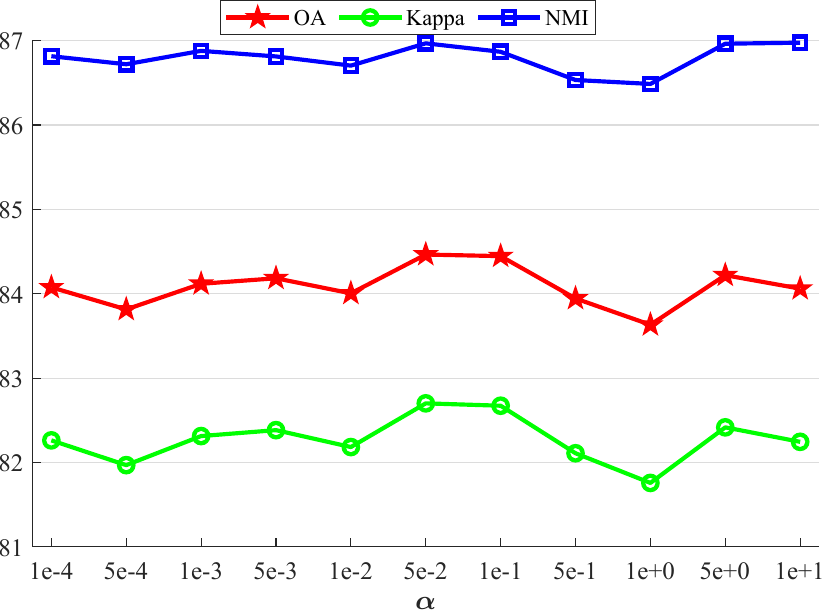}\label{SA_param_alpha}}
  \hfill
  \subfloat[]{\includegraphics[width=0.32\textwidth]{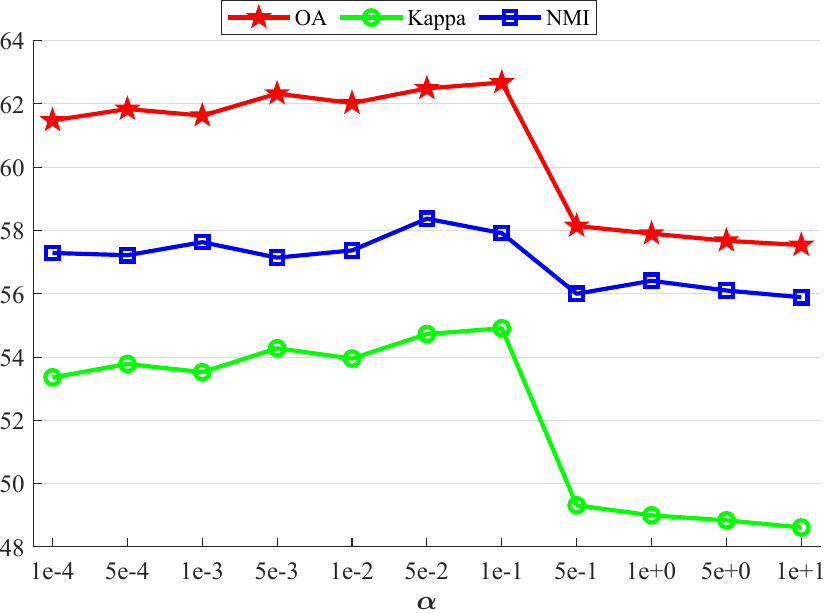}\label{PU_param_alpha}}
  \caption{The influence of different weight of clustering-center-level contrast loss $\alpha$ on (a) \textit{India Pines}, (b) \textit{Salinas}, (c) \textit{Pavia University}.}
  \label{param_alpha}
\end{figure*}

\begin{figure*}[ht]
  \centering
  \subfloat[]{\includegraphics[width=0.32\textwidth]{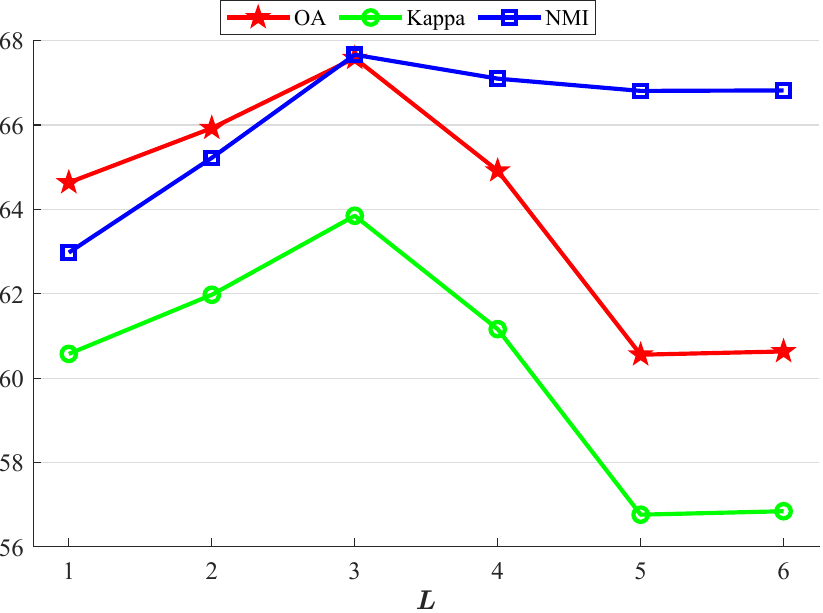}\label{IP_param_L}}
  \hfill
  \subfloat[]{\includegraphics[width=0.32\textwidth]{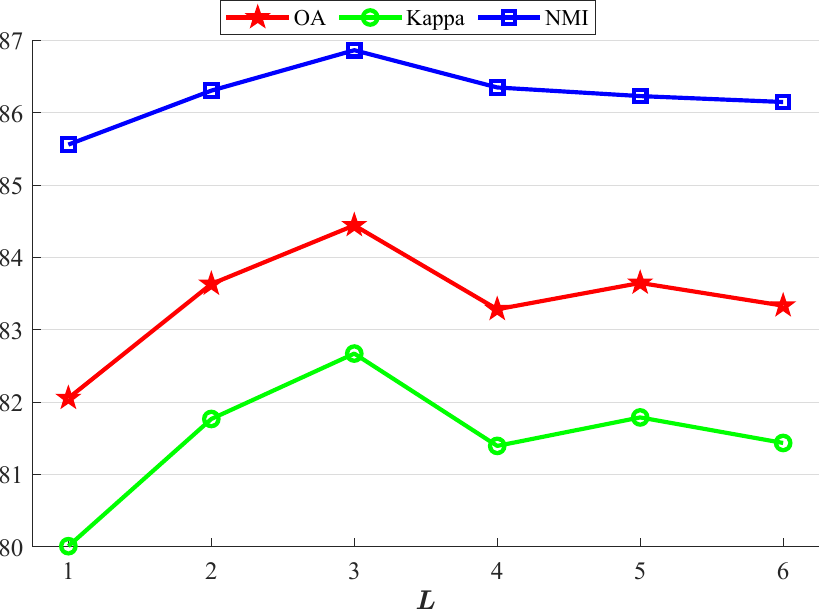}\label{SA_param_L}}
  \hfill
  \subfloat[]{\includegraphics[width=0.32\textwidth]{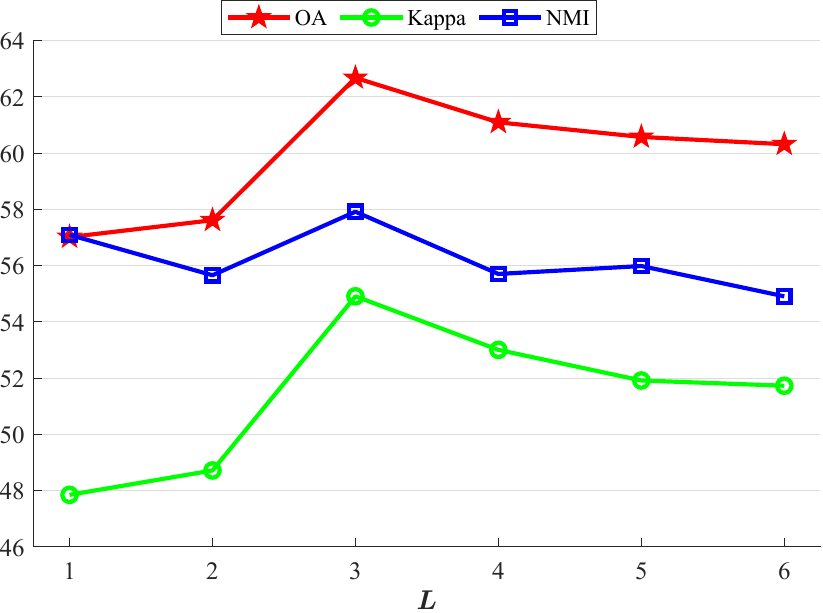}\label{PU_param_L}}
  \caption{The influence of different numbers of GCN layers $L$ on (a) \textit{India Pines}, (b) \textit{Salinas}, (c) \textit{Pavia University}.}
  \label{param_L}
\end{figure*}

\textit{2) Impact of $\lambda$:} $\lambda$ controls the ratio of selected samples when recomputing high-confidence clustering centers. According to Fig. \ref{param_lambda}, the proposed SPGCC is quite robust to parameter $\lambda$. Specifically, fluctuations of the clustering performance on the \textit{India Pines} dataset are slightly greater than those on \textit{Salinas} and \textit{Pavia University}. One possible reason is that the noise intensity on \textit{India Pines} is larger than that on other datasets and the number of divided superpixels is smaller, so a larger $\lambda$ makes clustering centers more susceptible to noise samples, while a smaller $\lambda$ means only a small part of samples being selected which cannot fully represent centers of certain classes. Tuning results show that the optimal values of $\lambda$ for the three datasets are respectively 0.75, 0.55, and 0.25.

\textit{3) Impact of $\alpha$:} $\alpha$ introduces the clustering-center-level contrast loss. Fig. \ref{param_alpha} suggests that as $\alpha$ increases from a very small value, the improvement brought by the clustering-center-level contrast gradually increases. When $\alpha$ is equal to 0.1, clustering performances on three datasets are almost optimal, although it is slightly better when $\alpha$ is set to 0.05 on \textit{Salinas}. When $\alpha$ exceeds 0.1, clustering performance declines significantly, indicating that a too-large weight for clustering-center-level contrast loss is harmful to superpixel representation learning. For the sake of simplicity, we finally fix $\alpha$ to 0.1.

\textit{4) Impact of $L$:} $L$ is the number of GCN layers. As shown in Fig. \ref{param_L}, it is evident that as $L$ increases, clustering metrics exhibit an increasing-then-decreasing trend. The best clustering performance is achieved when $L$ is set to 3 on all three datasets. Since one layer GCN can aggregate $1$-hop neighbors, small $L$  makes the model inaccessible to high-order neighbors, while large $L$ may lead to the over-smoothing problem, i.e., representations of different samples tend to be consistent and lack discriminability. Therefore, we empirically fix $L$ to 3.

\begin{table}[ht]
  \caption{Ablation Study Results, OA (\%) is Repored.\label{ablation_study_table}}
  \centering
  \resizebox{\linewidth}{!}{
    \begin{tblr}{
      row{1} = {c},
      column{2,3,4} = {c},
      vline{2} = {-}{0.03em},
      hline{1,9} = {-}{0.08em},
      hline{2,4,6,8} = {-}{0.03em},
        }
      Method                               & \textit{India Pines} & \textit{Salinas} & \textit{Pavia University} \\
      w/o pixel-level pre-training         & 53.54                & 67.83            & 50.49                     \\
      w/o high-confidence centers          & 65.48                & 83.64            & 62.18                     \\
      w/o pixel sampling augmentation      & 66.15                & 84.19            & 61.64                     \\
      w/o model weight augmentation        & 64.16                & 82.47            & 59.54                     \\
      w/o sample-level alignment           & 65.03                & 84.18            & 59.93                     \\
      w/o clustering-center-level contrast & 65.34                & 83.65            & 61.37                     \\
      SPGCC                                & \textbf{67.59}       & \textbf{84.47}   & \textbf{62.68}
    \end{tblr}}
\end{table}

\begin{table}[ht]
  \caption{Different Data Augmentations for Superpixels, OA (\%) is Repored, Pixel Sampling Augmentation and Model Weight Augmentation are Abbreviated to ``PSA" and ``MWA", Bold and Underline Denote the Optimal and Suboptimal Values Respectively.\label{sp_aug}}
  \resizebox{\linewidth}{!}{
    \centering
    \begin{tblr}{
      row{1} = {c},
      row{3} = {c},
      row{4} = {c},
      row{7} = {c},
      row{8} = {c},
      cell{2}{1} = {r=3}{},
      cell{2}{2} = {c},
      cell{2}{3} = {c},
      cell{2}{4} = {c},
      cell{2}{5} = {c},
      cell{5}{2} = {c},
      cell{5}{3} = {c},
      cell{5}{4} = {c},
      cell{5}{5} = {c},
      cell{6}{1} = {r=3}{},
      cell{6}{2} = {c},
      cell{6}{3} = {c},
      cell{6}{4} = {c},
      cell{6}{5} = {c},
      vline{2,3} = {-}{0.03em},
      hline{1,9} = {-}{0.08em},
      hline{2,5-6} = {-}{0.03em},
        }
      Type               & Augmentation      & \textit{India Pines} & \textit{Salinas} & \textit{Pavia University} \\
      Node-level         & Node mask         & 62.74                & 82.25            & 58.01                     \\
                         & Node noise        & 62.02                & 82.17            & 57.83                     \\
                         & Node shuffling    & 64.01                & 82.30            & 58.22                     \\
      Edge-level         & Edge perturbation & 63.27                & 83.77            & 58.99                     \\
      Semantic-invariant & PSA               & 64.16                & 82.47            & 59.54                     \\
                         & MWA               & \uline{66.15}        & \uline{84.19}    & \uline{61.64}             \\
                         & PSA+MWA           & \textbf{67.59}       & \textbf{84.47}   & \textbf{62.68}
    \end{tblr}}
\end{table}

\begin{figure*}[ht]
  \centering
  \subfloat[]{\includegraphics[width=0.32\textwidth]{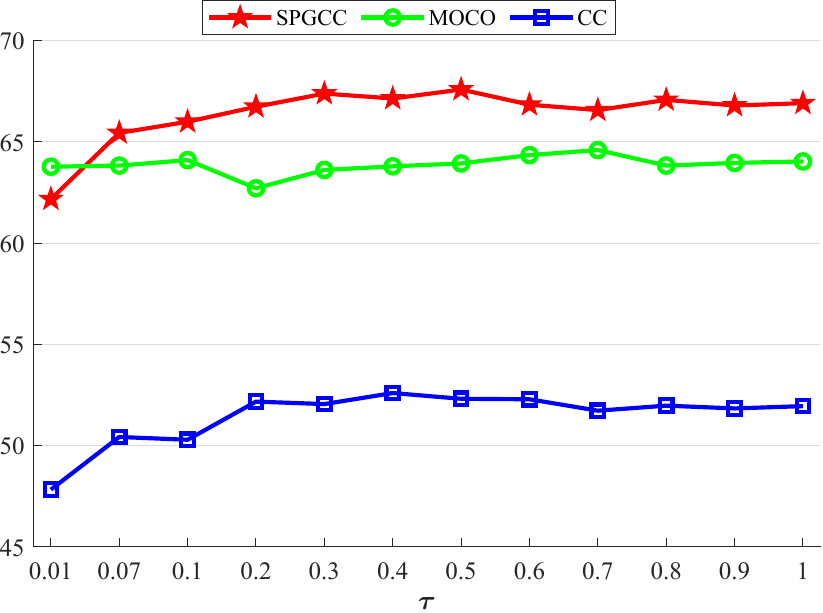}\label{IP_8}}
  \hfill
  \subfloat[]{\includegraphics[width=0.32\textwidth]{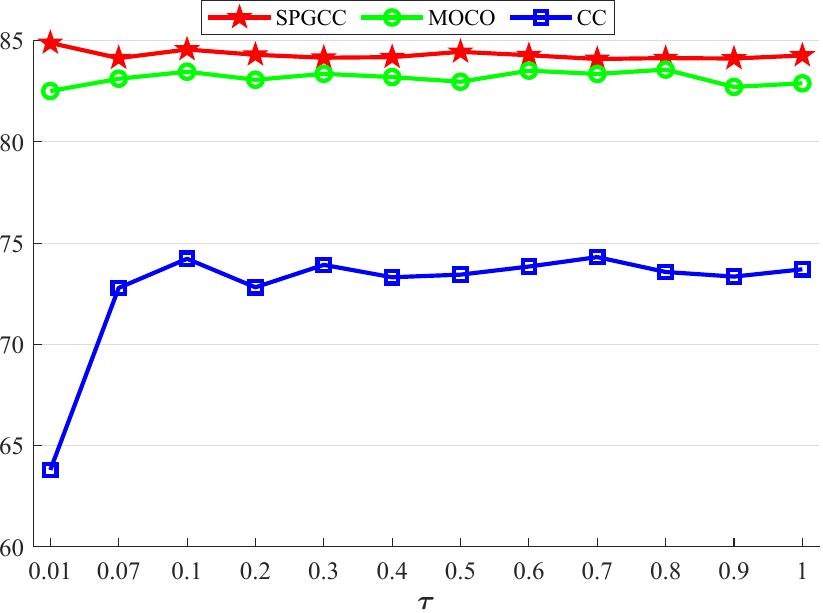}\label{SA_8}}
  \hfill
  \subfloat[]{\includegraphics[width=0.32\textwidth]{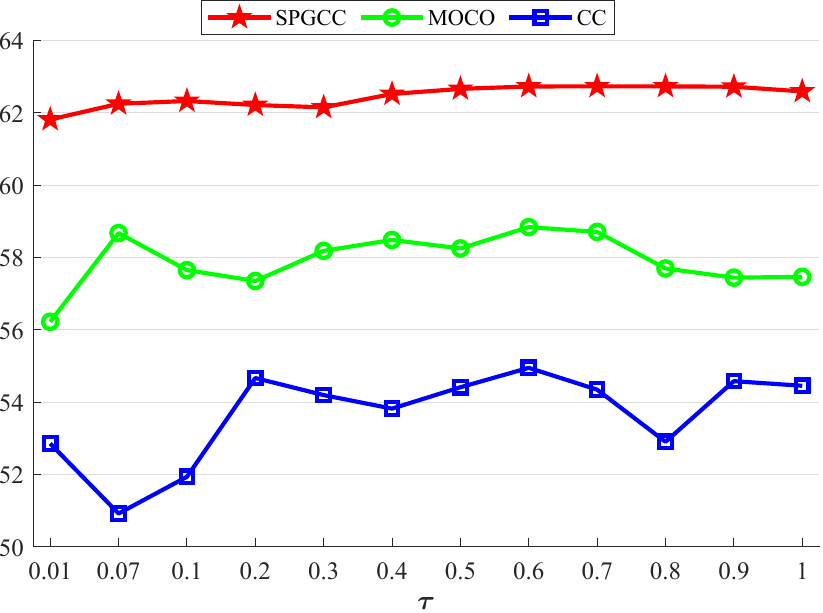}\label{PU_8}}
  \caption{OA (\%) obtained by different contrastive learning methods with different temperature scaling parameters $\tau$ on (a) \textit{India Pines}, (b) \textit{Salinas}, (c) \textit{Pavia University}.}
  \label{contrastive_learning_methods}
\end{figure*}

\begin{figure}[ht]
  \centering
  \includegraphics[width=\linewidth]{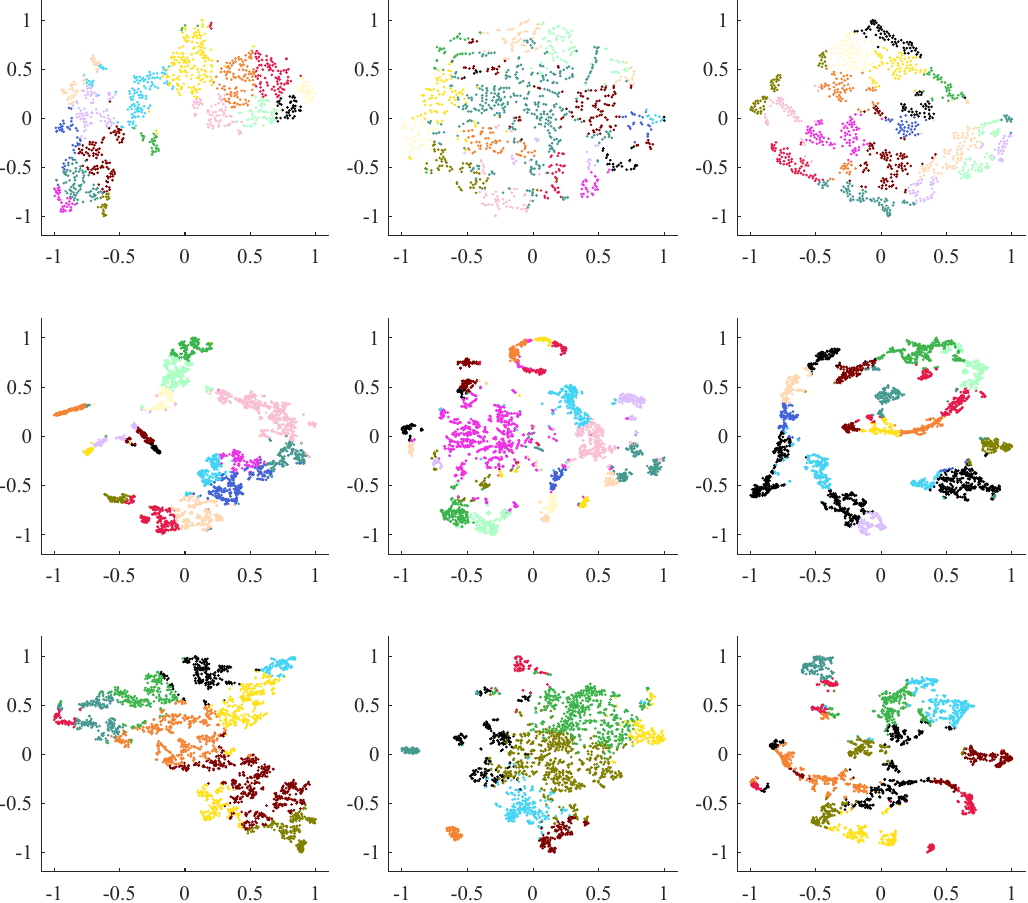}
  \caption{t-SNE visualization on \textit{India Pines} (first row), \textit{Salinas} (second row) and Pavia university (third row) by different features: raw features (first column), pixel-level pre-trained features (second column) and SPGCC learned features (third column).}
  \label{tsne}
\end{figure}

\subsection{Ablation Study\label{ablation_study}}
To validate the effectiveness of each module in the proposed SPGCC, including pixel-level pre-training, recomputed high-confidence clustering centers, two types of semantic-invariant superpixel augmentations (i.e., pixel sampling augmentation and model weight augmentation), two clustering-oriented training targets (i.e., sample-level alignment and clustering-center-level contrast), we conduct a series of ablation experiments. According to Table \ref{ablation_study_table}, the complete SPGCC achieves the best performance. In addition, we summarize the following four observations. First, pixel-level pre-training is very important and brings more than 10\% performance improvement, indicating that high-order spectral and spatial features extracted from pre-training are more conducive to clustering. Second, recomputing clustering centers only with a certain proportion of high-confidence samples can alleviate the influence of noise samples and thus improve clustering performance. Third, using two types of semantic-invariant superpixel data augmentations simultaneously performs better than only using one, which is consistent with the current opinion in self-supervised contrastive learning that multiple reasonable data augmentations are beneficial to representation learning. Fourth, combining sample-level alignment with clustering-center-level contrast performs better, indicating that sample-level alignment helps to increase intra-class similarity while clustering-center-level contrast helps to increase inter-class dissimilarity, thus making representations easier to cluster.

\subsection{Different Superpixel-level Data Augmentations}
We also compared the proposed two semantic-invariant superpixel augmentations, i.e., pixel sampling augmentation (PSA) and model weight augmentation (MWA), with general graph data augmentation methods\cite{Data_Augmentation_on_Graphs} which may hurt the node semantic in Table \ref{sp_aug}. The compared data augmentations include node mask, node noise, node shuffling, and edge perturbation. We tune the changing rate from 10\% to 30\% and report their best performance.

According to the experimental results in Table \ref{sp_aug}, the best clustering result is achieved by combining PSA and MWA. When only one type of augmentation is used, MWA outperforms the others, while PSA ranks second on \textit{India Pines} and \textit{Pavia University}, and third on \textit{Salinas}, showing the advantage of proposed semantic-invariant augmentations. In addition, general graph data augmentation methods require additional hyperparameters to control the intensity of augmentation, while PSA and MWA do not but still perform better, indicating the convenience and effectiveness of the designed augmentations.

\subsection{Different Contrastive Learning Methods}
Then, we compare the sample-level alignment (SLA) and clustering-center-level contrast (CLC) with other well-known contrastive learning methods, i.e., MOCO \cite{moco} and CC \cite{cc}. For fairness, we report the performance with different temperature scaling parameters $\tau$ for each method.

From Fig. \ref{contrastive_learning_methods}, it can be observed that both MOCO and CC are inferior to the proposed SPGCC. This validates the shortcoming of the sample-level contrast strategy adopted in MOCO and CC, which may treat samples in the same class as negative samples, pushing them away in the representation space and disrupting intra-class similarity. While our method avoids this problem by combining SLA and CLC, leading to better clustering performance.

\subsection{Visualization of Superpixel Representations with t-SNE}
$t$-distributed stochastic neighbor embedding (t-SNE)\cite{tsne_paper} is used for visualization of superpixel representations. We show the visualization results of the raw HSI features, pixel-level pre-trained features, and embeddings after SPGCC training in Fig. \ref{tsne}.

The raw features of superpixels are difficult to distinguish and are unevenly distributed in feature space. The clustering performance of the pixel-level pre-trained features has been improved, but since the target of the pre-training stage is not clustering-oriented, the distribution of superpixel representations is relatively loose. After SPGCC training, both intra-class similarity and inter-class dissimilarity of the features have been improved. It can be seen that the distribution of features in the same class is more compact, while features from different classes have better discriminability.

\section{Conclusion\label{conclusion}}
In this article, we proposed a superpixel graph contrastive clustering (SPGCC) algorithm to learn clustering-friendly representations of HSI. In SPGCC, high-order spatial and spectral information of HSI is extracted and preserved by pixel-level pre-training with 3-D and 2-D hybrid CNNs. To improve superpixel-level contrastive clustering, we design two semantic-invariant superpixel data augmentations, i.e., pixel sampling augmentation and model weight augmentation, where internal pixels and the output of dual-branch GCN serve as augmented views of superpixels. In addition, we recompute the high-confidence center of each class to mitigate the influence of outliers. Compared with general graph data augmentations, the proposed augmentations are more reliable. Clustering is guided by sample-level alignment and clustering-center-level contrast, which can overcome the weakness of common contrastive learning that pushes samples in the same class away and help learn the representations with intra-class similarity and inter-class dissimilarity. Experimental results demonstrated that SPGCC outperforms compared methods on different HSI datasets significantly. In the future, we hope to mine the more accurate correlation between superpixels to further guide superpixel-level graph learning for HSI.


\begin{IEEEbiography}[{\includegraphics[width=1in,height=1.25in,clip,keepaspectratio]{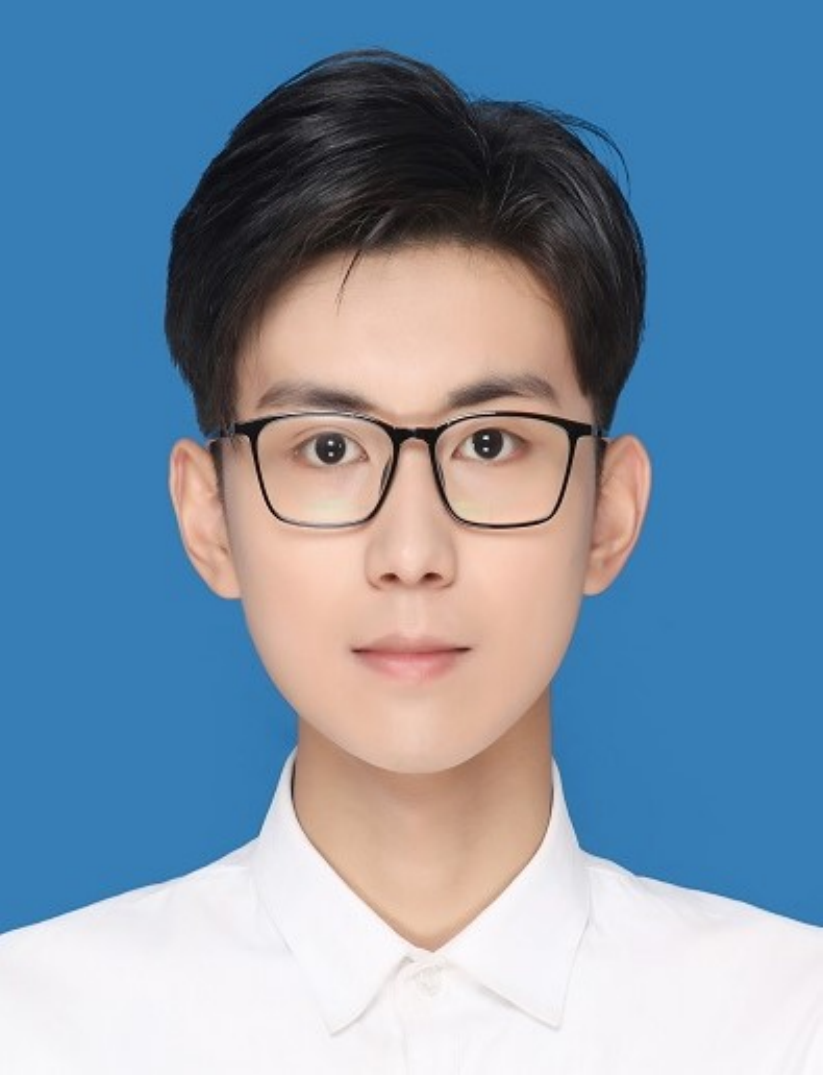}}]{Jianhan Qi}
  received the B.S. degree in Computer Science and Technology from Wuhan University of Technology, Wuhan, China in 2022. He is currently working toward the M.S. degree in Software Engineering from Southeast University, Nanjing, China. His research interests include unsupervised learning and graph neural networks.
\end{IEEEbiography}

\begin{IEEEbiography}[{\includegraphics[width=1in,height=1.25in,clip,keepaspectratio]{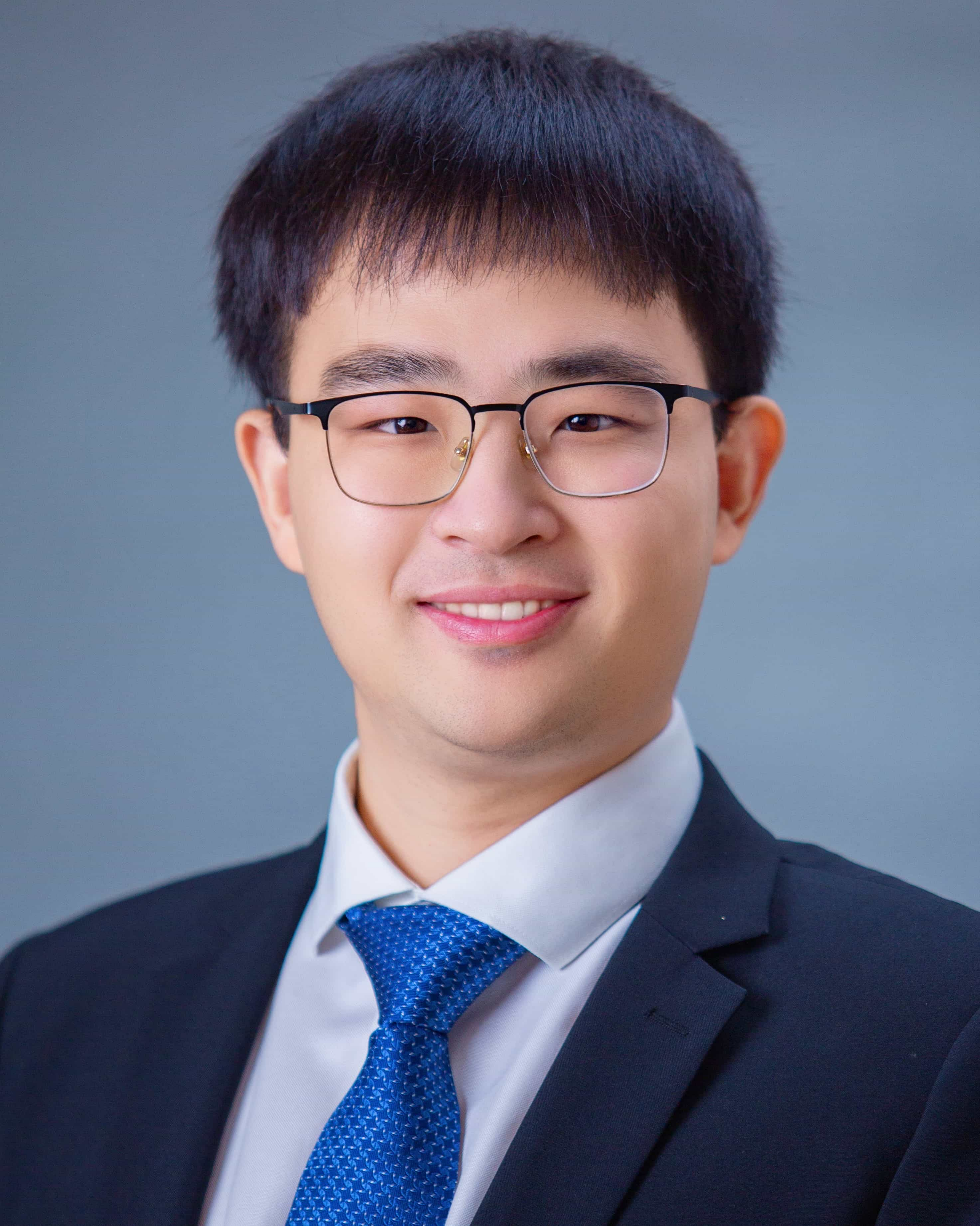}}]{Yuheng Jia}
  (Member, IEEE) received the B.S. degree in automation and the M.S. degree in control theory and engineering from Zhengzhou University, Zhengzhou, China, in 2012 and 2015, respectively, and the Ph.D. degree in computer science from the City University of Hong Kong, Hong Kong, China, in 2019.

  He is currently an Associate Professor with the School of Computer Science and Engineering, Southeast University, Nanjing, China. His current research interests include machine learning and data representation, such as weakly-supervised learning, high-dimensional data modeling and analysis, and low-rank tensor/matrix approximation and factorization.
\end{IEEEbiography}

\begin{IEEEbiography}[{\includegraphics[width=1in,height=1.25in,clip,keepaspectratio]{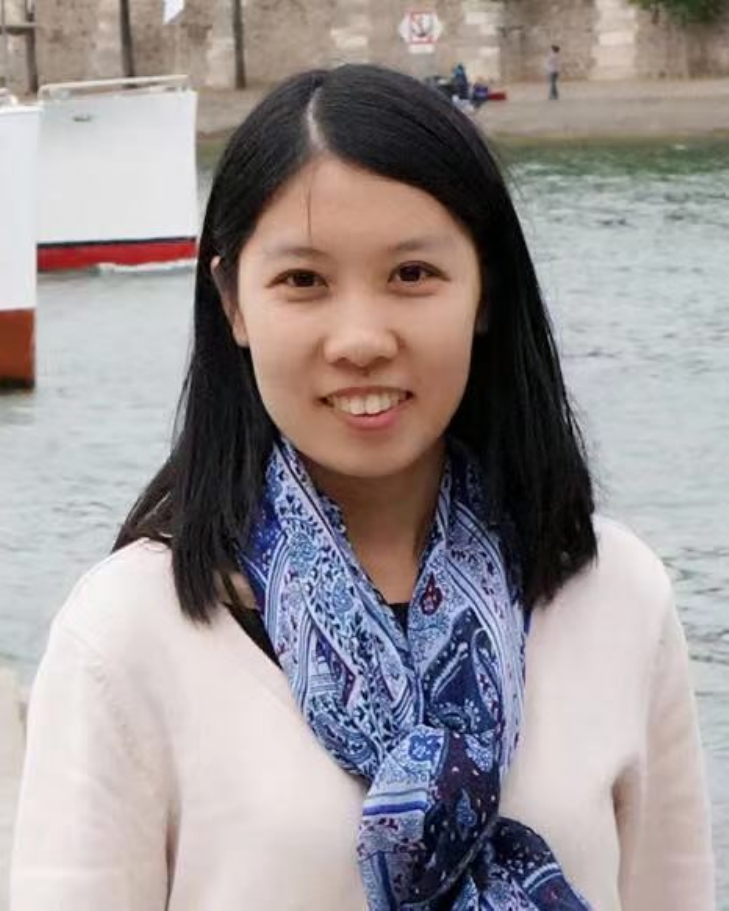}}]{Hui Liu}
  received the B.Sc. degree in communication engineering from Central South University, Changsha, China, the M.Eng. degree in computer science from Nanyang Technological University, Singapore, and the Ph.D. degree in computer science from the City University of Hong Kong, Hong Kong, China. From 2014 to 2017, she was a Research Associate with the Maritime Institute, Nanyang Technological University. She is currently an Assistant Professor with the Department of Computing and Information Sciences, Saint Francis University, Hong Kong, China. Her current research interests include image processing and machine learning.
\end{IEEEbiography}

\begin{IEEEbiography}[{\includegraphics[width=1in,height=1.25in,clip,keepaspectratio]{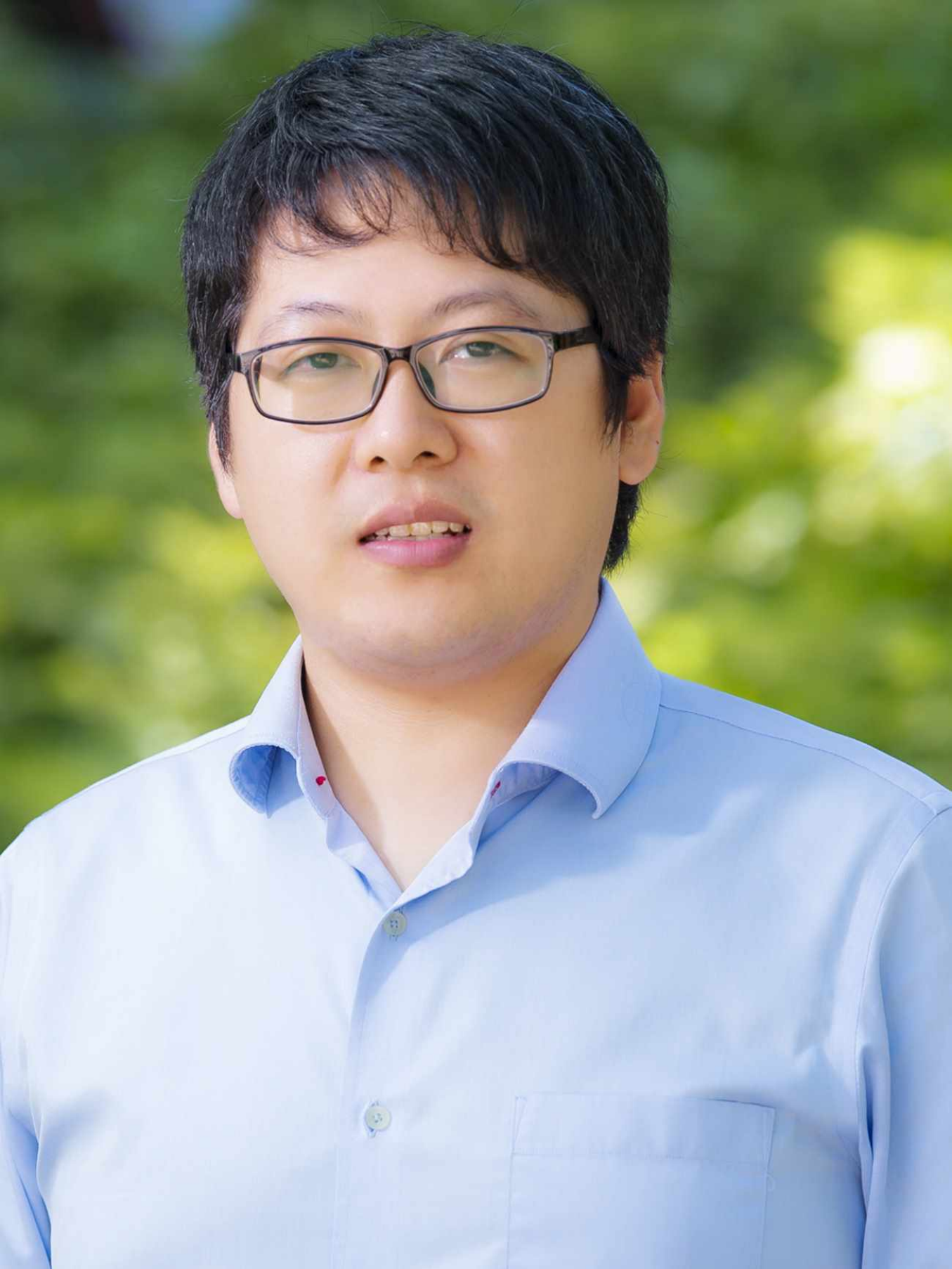}}]{Junhui Hou}
  (Senior Member, IEEE) received the B.Eng. degree in information engineering (talented students program) from the South China University of Technology, Guangzhou, China, in 2009, the M.Eng. degree in signal and information processing from Northwestern Polytechnical University, Xi’an, China, in 2012, and the Ph.D. degree from the School of Electrical and Electronic Engineering, Nanyang Technological University, Singapore, in 2016.

  He is currently an Associate Professor with the Department of Computer Science, City University of Hong Kong. His current research interests include multi-dimensional visual computing. He is an Elected Member of IEEE MSA-TC, VSPC-TC, and MMSP-TC. He received the Early Career Award (3/381) from the Hong Kong Research Grants Council in 2018. He is currently serving as an Associate Editor for IEEE Transactions on Visualization and Computer Graphics, IEEE Transactions on Circuits and Systems for Video Technology, IEEE Transactions on Image Processing, Signal Processing: Image Communication, and The Visual Computer.
\end{IEEEbiography}

\vfill
\end{document}